\RequirePackage{fix-cm}
\documentclass[natbib,twocolumn]{svjour3}          
\smartqed  
\usepackage{pifont}
\usepackage{graphicx}
\usepackage{float}
\usepackage{subfig}
\usepackage{microtype}
\usepackage{xspace}
\usepackage{algorithm}
\usepackage{algorithmicx}
\usepackage{algpseudocode}

\usepackage{booktabs}
\usepackage{tabularx}
\newcolumntype{Y}{>{\centering\arraybackslash}X}
\usepackage{xspace}
\usepackage{esvect}
\usepackage{xcolor}
\usepackage[pagebackref=true,breaklinks=true,citecolor=tealblue,colorlinks,linkcolor=ruby,bookmarks=false]{hyperref}
\definecolor{ruby}{rgb}{0.88, 0.07, 0.37}
\definecolor{tealblue}{rgb}{0.18, 0.40, 0.46}
\usepackage{array}
\usepackage{multirow}
\newcolumntype{H}{>{\setbox0=\hbox\bgroup}c<{\egroup}@{}}
\usepackage{amsmath}
\usepackage{amsfonts}
\usepackage{bm}
\usepackage{bbm}
\usepackage{mathrsfs}
\usepackage{mathrsfs}

\begin{document}
	\title{{A Dimensional Structure based Knowledge Distillation Method for Cross-Modal Learning}}
	\author{Lingyu~Si    \and
		Hongwei~Dong    \and
		Wenwen~Qiang            \and
		Junzhi~Yu         \and
		Wenlong~Zhai         \and
		Changwen~Zheng         \and
		Fanjiang~Xu         \and   Fuchun~Sun
	}
	\institute{Lingyu Si, Hongwei Dong, Wenwen Qiang, Wenlong Zhai, Changwen Zheng and Fanjiang Xu \at
		Science \& Technology on Integrated Information System Laboratory, Institute of Software Chinese Academy of Sciences\\
		University of Chinese Academy of Sciences\\
		\email{\{lingyu, donghongwei, qiangwenwen, zhaiwenlong, changwen, fanjiang\}@iscas.ac.cn}          
		\and
		Junzhi Yu \at
		College of Engineering, Peking University\\
		\email{yujunzhi@pku.edu.cn}
		Fuchun Sun \at
		Department of Computer Science and Technology, Tsinghua University\\
		\email{fcsun@mail.tsinghua.edu.cn}
		\and
		Lingyu Si and Hongwei Dong contributed equally to this work 
		\\
		Corresponding author: Wenwen Qiang
	}
	\date{Received: date / Accepted: date}
	
	\def\ourconv{RIConv++\xspace}
	\def\smallgap{\vspace{0.05in}}
	\maketitle
	\begin{abstract}
		Due to limitations in data quality, some essential visual tasks are difficult to perform independently. Introducing previously unavailable information to transfer informative dark knowledge has been a common way to solve such hard tasks. However, research on why transferred knowledge works has not been extensively explored. To address this issue, in this paper, we discover the correlation between feature discriminability and dimensional structure (DS) by analyzing and observing features extracted from simple and hard tasks. On this basis, we express DS using deep channel-wise correlation and intermediate spatial distribution, and propose a novel cross-modal knowledge distillation (CMKD) method for better supervised cross-modal learning (CML) performance. The proposed method enforces output features to be channel-wise independent and intermediate ones to be uniformly distributed, thereby learning semantically irrelevant features from the hard task to boost its accuracy. This is especially useful in specific applications where the performance gap between dual modalities is relatively large. Furthermore, we collect a real-world CML dataset to promote community development. The dataset contains more than 10,000 paired optical and radar images and is continuously being updated. Experimental results on real-world and benchmark datasets validate the effectiveness of the proposed method.
	\end{abstract}
	
	\section{Introduction}\label{sec:introduction}
\begin{figure*}[ht]
		\centering
		\includegraphics[width=0.886\textwidth]{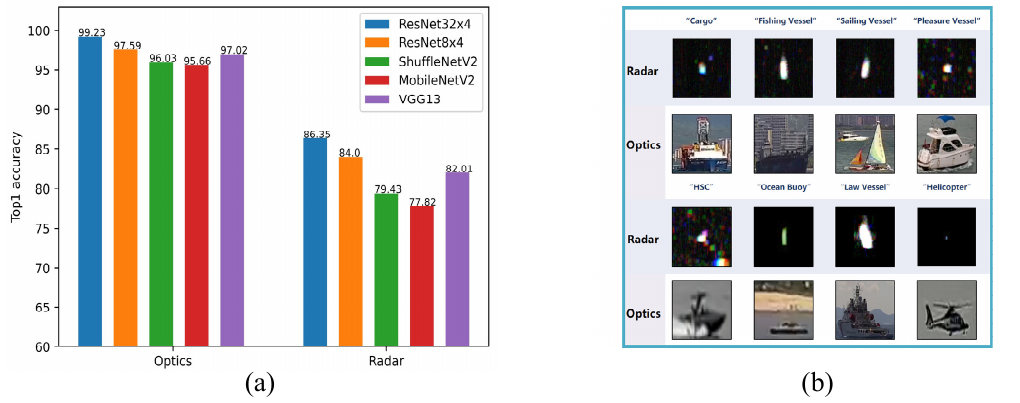}
		\caption{A simple test on the difficulty of identifying the same object under different modalities. Limited by data quality, the recognition accuracy for radar images is significantly lower than that for optical images. (a) Results of classification accuracy. (b) Quality comparison of paired cross-modal data.}
		\label{fig:motivate1}
	\end{figure*}

	The success of deep learning has significantly promoted the development of real-world applications \citep{Kim2016eccv, Hu2020eccv,9165231,9351574}. Nevertheless, there remain numerous tasks that are difficult to resolve using deep learning-based methods. Part of this problem can be attributed to the lack of customized algorithms, but the primary reason is the inadequacy of data quality. To exemplify this point, let us consider the simple example illustrated in Fig. \ref{fig:motivate1}, where we assess the accuracy of several extensively utilized deep image classifiers \citep{vggnet,inceptionv4,7780459,8578843,huang2017densely} in recognizing identical objects across different visual modalities (for more information, please refer to Subsection \ref{subsec:motiexp}). The results demonstrate that data quality has a far greater impact on classification accuracy than algorithm design.

	Based on the observation of the motivating example mentioned above, a straightforward solution to enhance certain hard visual tasks in modalities with relatively low data quality is to utilize other informative signals, if available, to import the key information required for accurate prediction. This solution falls into the area of cross-modal learning (CML). CML aims to model the relationship between dual data modalities to leverage multimodal information and improve performance in the ensemble or single manners \citep{Liu2020}. In recent research, multi-stream networks \citep{8447210}, attention mechanisms \citep{KimandChoi2018eccv, Xu2016eccv}, and adversarial learning \citep{8974594,9174895} are common techniques employed for addressing diverse tasks in CML, such as data classification \citep{9259207}, image understanding \citep{9292444}, retrieval \citep{8563047}, and so forth.
	\begin{figure*}[!ht]
		\centering
		\subfloat[][]{\includegraphics[width=0.24\textwidth]{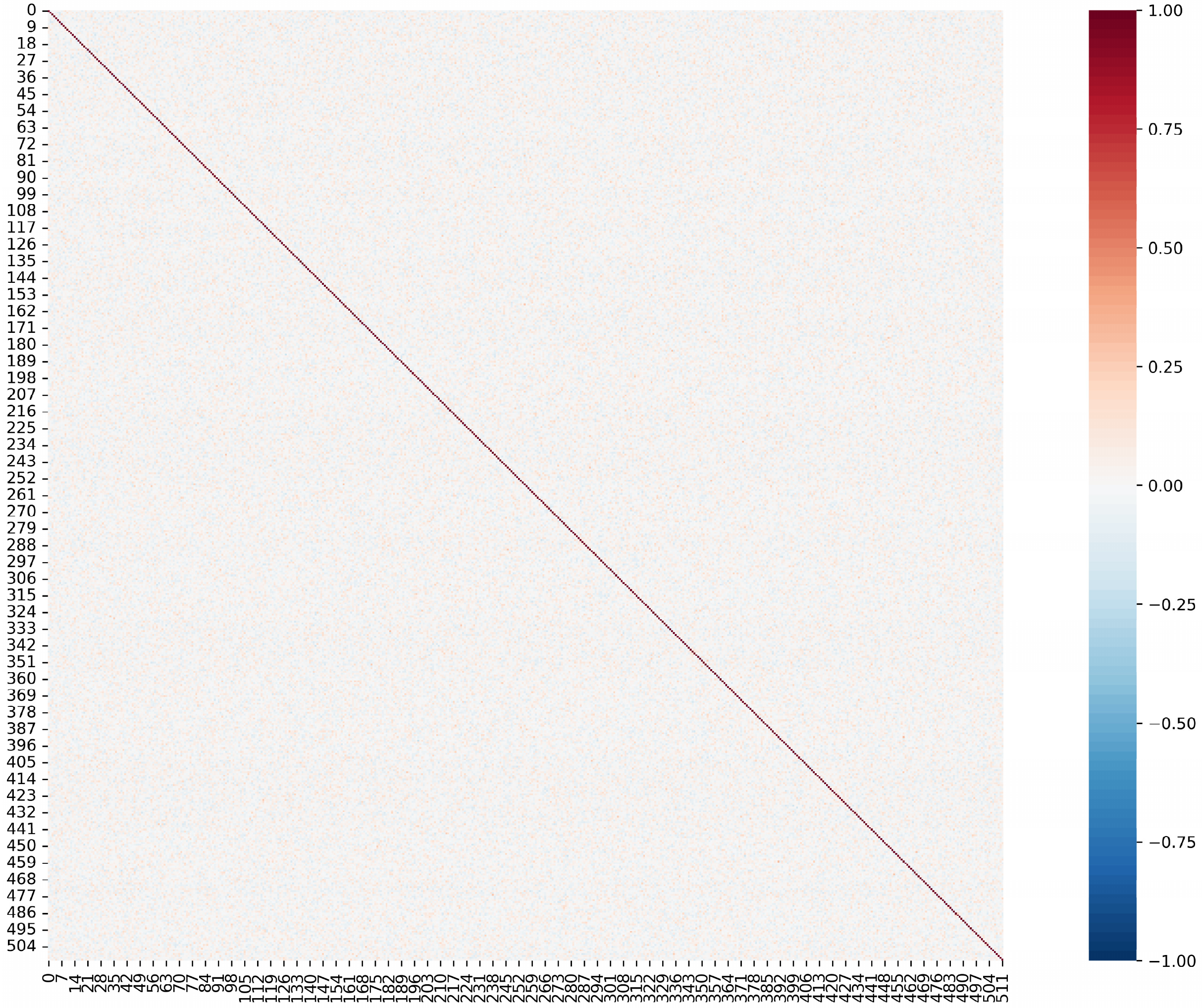}\label{fig:motivate2a}}
		\hspace{0.2em}
		\subfloat[][]{\includegraphics[width=0.24\textwidth]{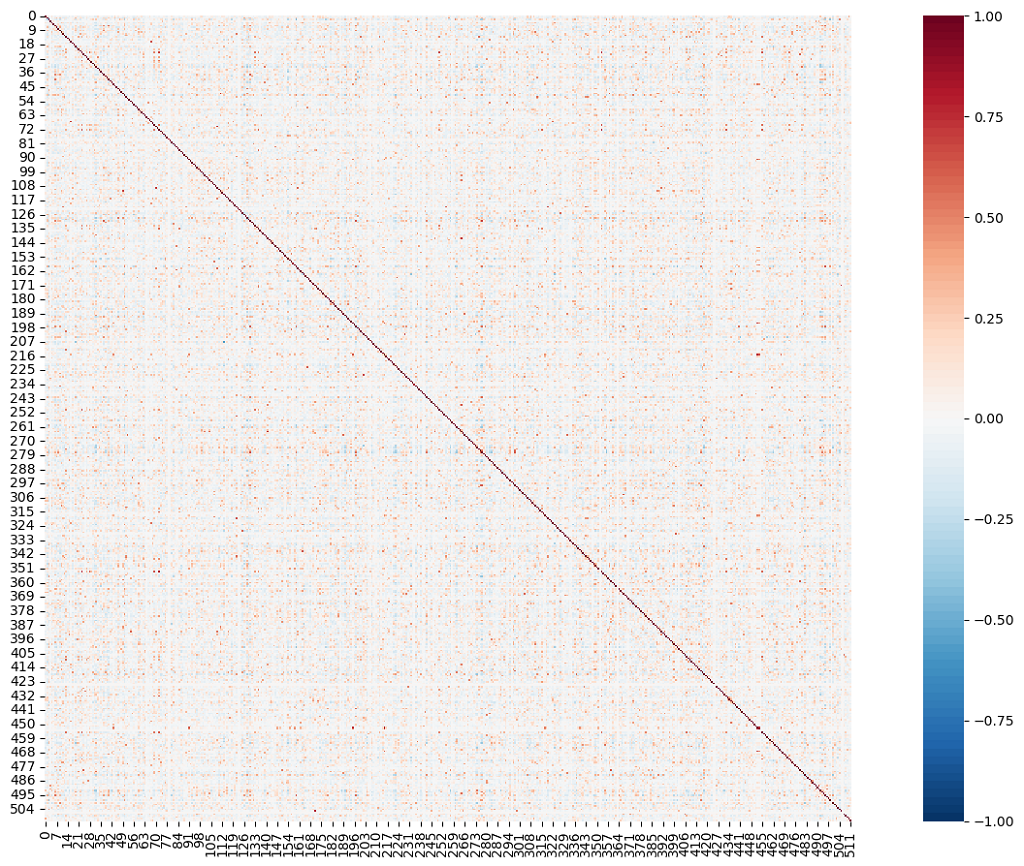}\label{fig:motivate2b}} 
		\hspace{0.2em}
		\subfloat[][]{\includegraphics[width=0.24\textwidth]{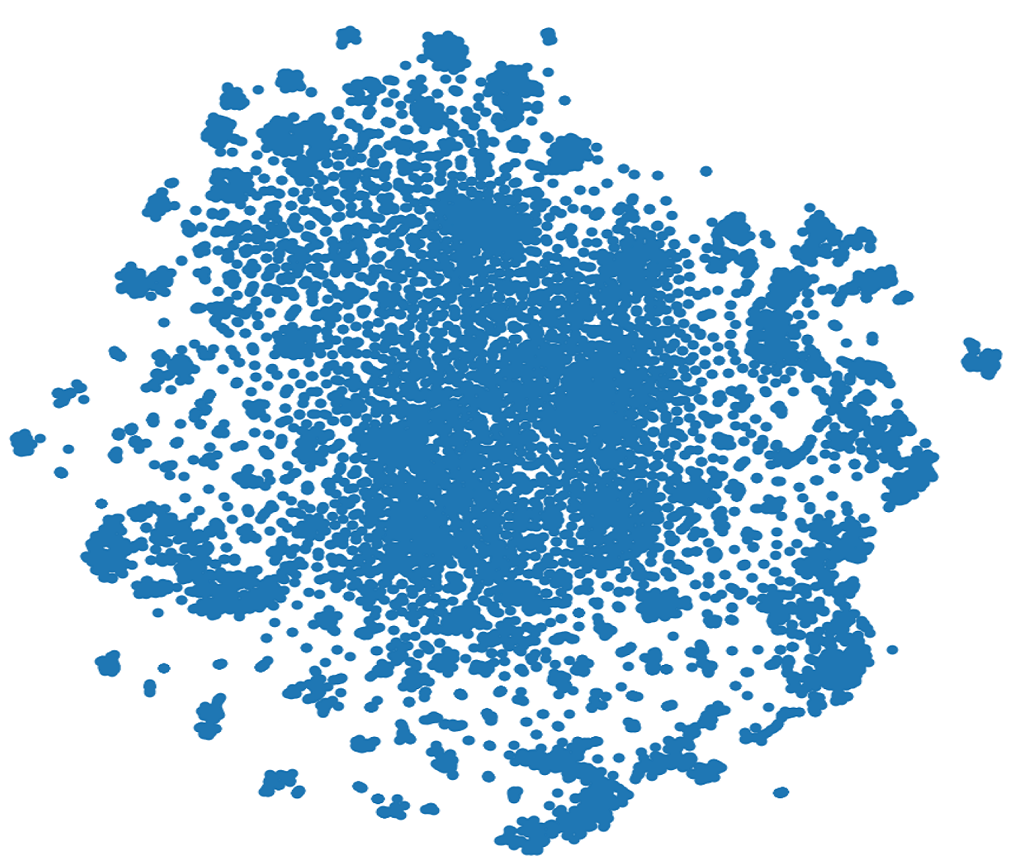}\label{fig:motivate2c}}
		\hspace{0.2em}
		\subfloat[][]{\includegraphics[width=0.24\textwidth]{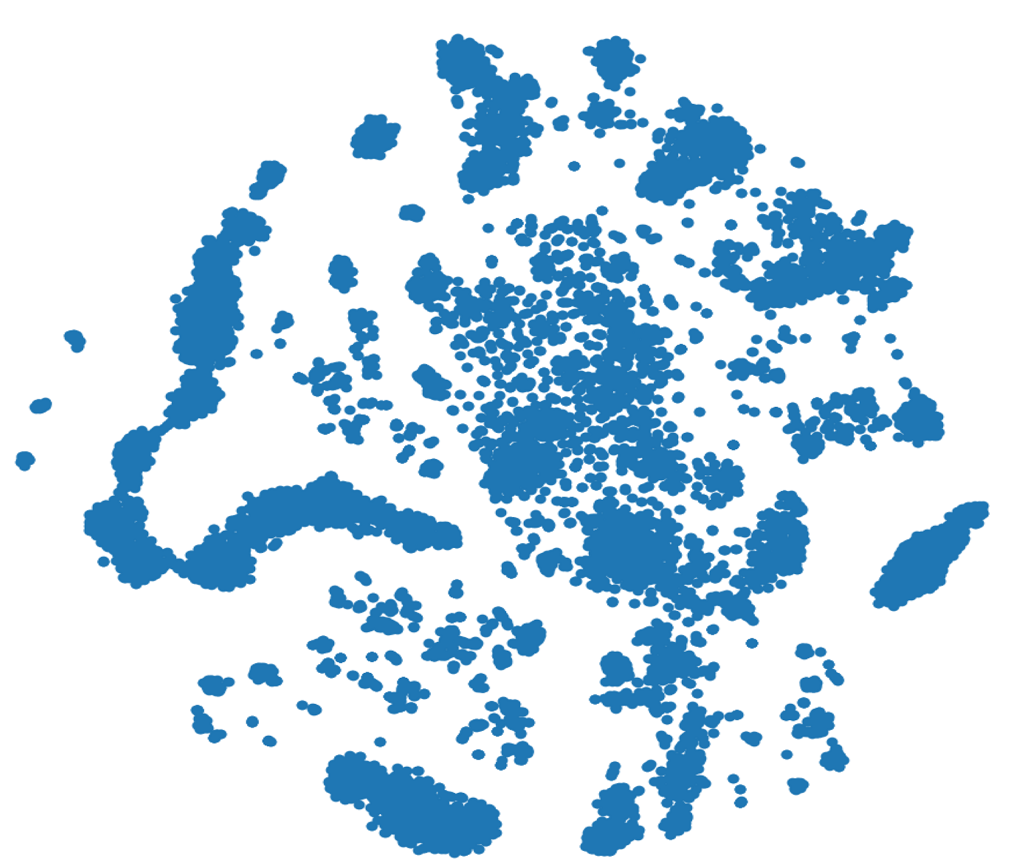}\label{fig:motivate2d}}    
		\caption{Visualizations of DS from two different aspects. (a)-(b) Self-similarity matrices of output features extracted from optics and radar, respectively. (c)-(d) Spatial distributions of intermediate features extracted from optics and radar, respectively.}
		\label{fig:motivate2}
	\end{figure*}
	
	Knowledge transfer plays a pivotal role in CML and is often accomplished through knowledge distillation (KD) \citep{gou2021knowledge}. The seminal work of KD was proposed by Hinton \textit{et al.}, which achieved model compression on homologous data by compelling a lightweight student network to mimic the output logits of a cumbersome teacher network \citep{HintonKD}. In subsequent works, diverse types of intermediate representations \citep{romero2014fitnets,10.1007/978-3-030-58595-2_40,chen2021crossSEMCKD} or instance relations \citep{8100237FSP,Tung_2019_ICCV,park2019relational} have been used as surrogates for logits to attain better student accuracy. Furthermore, when the input data is heterogeneous, the primary objective of KD shifts from model compression to enhancing accuracy through knowledge transfer, known as cross-modal knowledge distillation (CMKD), and numerous practical application-oriented works have been investigated\citep{Liu_2021_CVPR,8237837,NEURIPS2020_6f2268bd,7487708,9357413}. However, most of these works focus on methodological enhancements to facilitate the better transfer of informative dark knowledge, and an analysis of why transferred knowledge works are lacking \citep{Tian2020CRD}.

	To address this issue, we decouple the process of knowledge transfer in CMKD and clarify the specific types of knowledge that should be transferred, based on the understanding of a question derived from the previous motivating example:
	\begin{itemize}
		\item[-] \textit{Compared with the hard task (radar image classification in Fig. \ref{fig:motivate1}), in what ways do features extracted from the simple task (optics) show their advantages?} 
	\end{itemize}
	
	Considering that each feature dimension corresponds to a kind of semantic information \citep{zeiler2014visualizing,zhou2016learning}, an intuitive conjecture for this question is that the discriminative features should have a better dimensional structure (DS). For instance, an ideal state can be characterized by a low-level correlation between each feature dimension. This implies that semantically irrelevant information \citep{barlow1961possible,pmlr-v89-esmaeili19a} has been learned, which is undoubtedly advantageous for enhancing feature discriminability.
	
	To evaluate this conjecture, we conduct a comparison between the DS, i.e., channel-wise self-similarity matrices of output features, obtained from simple and hard tasks in the aforementioned motivating example. The experimental results are depicted in Fig. \ref{fig:motivate2}(a)-(b), which provides empirical evidence that optics have a better DS compared to radar (for more details, kindly refer to Subsection \ref{subsec:motiexp}).
	
	Above, we take the output feature as the carrier of DS and make qualitative and quantitative analyses. However, as feature learning is a layer-by-layer process, there should also be some concept in the intermediate network layer that reflects the DS. For that matter, the distribution of intermediate features may be a suitable candidate, which manifests the richness of low-level visual information. When the intermediate features are uniformly distributed in their space, the information is maximally preserved, providing a basis for learning rich semantic information \citep{Jaynes1957}. To further explore this, we compare the distributions of intermediate features between the simple and hard tasks in the previous motivating example. Fig. \ref{fig:motivate2}(c)-(d) show that the distribution of optics is closer to uniform than that of radar (for more details, see Subsection \ref{subsec:motiexp}).
	
	Based on the insights and explanations outlined above, this work aims to improve the performance of hard tasks through the enhancement of DS. This is accomplished by explicitly transferring the structural knowledge contained in intermediate and output features. However, it is well-known that the development of related algorithms is heavily reliant on benchmark datasets \citep{WANGREVIEW}. To address this issue, many researchers have devoted their efforts to data collection \citep{Silberman2012, Liu_2021_CVPR,8237837,9844015,8758197}. But collecting paired cross-modal data remains challenging due to the need for substantial human and material resources. Consequently, many studies focus on the design of weakly-supervised CML methods using transfer learning \citep{pan2010survey,9236655}, adversarial learning \citep{goodfellow2020generative,9057710}, and self-supervised learning \citep{chen2020simple, NEURIPS2020_6f2268bd}. Although weakly-supervised algorithms alleviate the need for a large number of paired labeled data, their performance is inevitably degraded. In light of the aforementioned problems, we collect a real-world CML dataset in this work, which contains over 10,000 paired images of 8 different types of targets captured simultaneously by optical and radar sensors. The dataset is labeled, open source, and updated continuously.   
	
	Our contributions can be summarized as follows:
	\begin{enumerate}
		\item We investigate the influence of DS on feature discriminability and define the structural knowledge in terms of channel-wise correlation and spatial distribution. This enables us to introduce novel approaches by directly integrating DS-based methods and existing ones.
		\item We propose a novel DS-based method for CMKD that implements knowledge transfer by enforcing output features to transition from channel-wise correlation to independence, and intermediate features to be uniformly distributed in their space. These two aspects ensure that more semantic information can be learned from the modality with lower data quality.
		\item We collect and open source a CML dataset consisting of optical and radar images from real-world scenarios. This dataset is expected to be valuable for the advancement of the research community in the future.
		\item Experimental results, both on real-world and benchmark datasets, demonstrate that the proposed method outperforms prior state-of-the-art methods.
	\end{enumerate}
	
	\section{Related Works}
	\subsection{Cross-Modal Learning (CML)}
	CML is an active topic for application-oriented research that seeks to benefit from the powerful model in one modality to improve the less effective model in the other modality. Representative tasks of CML are recognition, understanding, and retrieval, where the features required for their implementation have different characteristics. Recognition is the most widely studied visual task, including classification, segmentation, detection, action recognition, etc. Extracting features with discriminability is the primary concern of such tasks. In contrast, the remaining two tasks are more challenging, such as visual question answering, image captioning, and text-image retrieval. These tasks aim to eliminate the heterogeneity between modalities and find a latent feature space where representations from dual modalities can be compared directly. 
	
	This work is concerned with CML for the recognition task. The requirement for the algorithm design is to handle the relationship between dual modalities reasonably and learn how to effectively use their information to obtain higher accuracy \citep{10.1007/978-3-031-19836-6_12}. Two different kinds of solutions have been developed to address this problem. One is to learn the modality-specific information and develop methods to extract useful and filter harmful information during the prediction process. Therefore, features of the two modalities present a complementary state. \citet{NEURIPS2019_d790c9e6} proposed an adaptive information combination method for cross-modal few-shot learning. \citet{Wang2022MM} used the random network prediction to estimate uncertainty and to combine cross-modal information. The other solution is to use the powerful model learned from one modality to help the learning on the other modality to reduce their performance gap. Sophisticated design to ensure knowledge transfer from the powerful model to the less effective model is critical to the success of such approaches. \citet{Afham_2022_CVPR} used the knowledge in 2D images to assist the learning of 3D point cloud data understanding task. \citet{Li_Wei_Hong_Gong_2020} introduced an auxiliary modality as an assistant to reduce the performance gap between infrared and optical learning tasks. Similarly, \citet{Ren_2021_CVPR} proposed a method of using audio data to assist the learning of silent lip videos, which improved the ways of knowledge acquisition and transfer. 
	
	It is worth noting that the above two solutions have their scope: When the performance gap between the modalities is small, learning complementary information is feasible. Otherwise, learning valuable information from the modality with lower data quality to benefit the better one may be difficult, and paying attention to knowledge transfer is likely to be more efficient. The former tends to make extensive use of the information of each modality. The latter focuses on improving the performance of the specific modality, which is also the interest of our concern. Unlike the existing methods, we propose a novel CMKD method based on transferring the structural knowledge contained in DS, which performs cross-modal updating on intermediate and output features to improve the learning performance.
	
	\subsection{Knowledge Distillation (KD)}
	Knowledge distillation (KD) \citep{gou2021knowledge}, also called the student-teacher learning paradigm, has been regarded as a mainstream method for model compression \citep{8253600}. \citet{HintonKD} proposed the seminal work of KD, where a trained powerful but cumbersome network model is called teacher, and a random-initialized lightweight model is called student. The teacher provides supervision for the student's training, so the model with slightly lower performance but significantly reduced model size can be obtained, which is helpful for edge devices. In early studies, taking the output as the knowledge to regress the teacher logits \citep{9156610}, and taking the arbitrary feature map as the knowledge to minimize the similarity between teacher and student \citep{Huang2017arxiv}, are the two main ways of KD for model compression. Later, many works use relationships between different layers or instances as the knowledge to be transferred, which shows strong competitiveness \citep{Tung_2019_ICCV,Tian2020CRD}. 
	
	Recently, some researchers have explored the potential of KD in CML \citep{9054253,9414480}. Compared with the KD for model compression, CMKD still uses information from the teacher as the supervisor to assist in the training of students \citep{7780678}, but the two models have comparable sizes and different input modalities. Therefore, CMKD pays more attention to improving the accuracy of the student by transferring some kind of informative dark knowledge.
	
	remote sensing image classification \citep{9772621}, text-image retrieval \citep{9945996}, thermal infrared tracking \citep{10.1145/3474085.3475387,9894056}, etc. At the same time, some other studies have been performed to boost the algorithm performance, including the modifications of loss function \citep{8014768,9577898}, the model architecture \citep{9879338}, and the dependence of annotation \citep{9879136}. The above-mentioned methods regard the logits, features, instance relations, or task-specific information as the knowledge to be transferred, and make corresponding methodological improvements. However, the question of why the transferred knowledge is effective and in which form it affects the discriminability of student features has not been answered. In this work, we provide an intuitive conjecture to this question through a combination of quantitative and qualitative analyses.
	
	\section{Preliminary Study and Analysis}
	To achieve the goal of using one modality to assist the learning of another, it is crucial to determine what kind of knowledge should be transferred. In our approach, we focus on transferring the structural knowledge contained in DS from both intermediate and output perspectives. Our motivation and analysis for this approach are detailed below.
	
	\subsection{Motivating Examples}\label{subsec:motiexp}
	Our motivating examples include two parts. First, we carry out experiments to investigate the difficulty of image classification under different data modalities. Specifically, we employ five popular deep learning-based classification methods, i.e., VGGNet \citep{vggnet}, ShuffleNet \citep{ma2018shufflenet}, MobileNet \citep{sandler2018mobilenetv2}, and two architectures of ResNet \citep{7780459}, to classify paired optical and radar images individually. Detailed experimental settings are described in Section \ref{sec:exp}.
	
	The results, as shown in Fig. \ref{fig:motivate1}(a), reveal a significant performance disparity when utilizing different data modalities. Notably, the optical image classification accuracy is remarkably high, indicating that the learned features are comprehensive. Conversely, the performance of radar image classification is unsatisfactory, suggesting a lack of sufficient discriminability in its features. Considering that each sensor has unique advantages, improving the learning performance of the hard task, e.g., radar classification in Fig. \ref{fig:motivate1}, with the help of the powerful model learned from the simpler task, e.g., optics, is a critical problem to solve.
	
	CMKD is a promising approach to facilitate learning in such hard tasks. However, the core issue lies in how to express the knowledge to be transferred, which brings us to the second part of our motivating examples. We test two critical concepts, i.e., channel-wise correlation and spatial distribution, of the output and intermediate features extracted from the trained models in the first motivating example. 
	Implementation details are as follows: First, we calculate the cosine similarity between the output feature and its transposition to evaluate channel-wise self-similarity. Next, we use t-SNE \citep{van2008visualizing} to reduce the dimension of intermediate features and visualize their distribution.
	
	The two tested concepts are expressions of DS from the perspective of output and intermediate features, respectively. We test them to verify an intuitive conjecture that features extracted from optical images have better DS compared to radar. This is reflected in lower channel-wise correlation and a more uniform spatial distribution. The results are shown in Fig. \ref{fig:motivate2}, which demonstrates the following:
	\begin{itemize}
		\item[-] The self-similarity matrices of output features extracted from optical images have off-diagonal elements with values closer to $0$ than those of radar images, empirically proving that better output features have lower channel-wise correlations.
		\item[-] The distributions of intermediate features of optical images have shorter spatial distances between each instance compared to radar, empirically verifying that better intermediate features distribute more uniformly.
	\end{itemize}

	\subsection{Problem Analysis}
	The research problem addressed in this work is initially identified while attempting to solving a practical application problem: Given the availability of paired optical and radar data, how to enhance the classification accuracy of radar images by leveraging optics? It is well-known that radar plays a crucial role in observation systems because of its ability to monitor in all-day and all-weather conditions, albeit with relatively lower data quality. In contrast, optics, which is commonly used in visual applications, boasts higher resolution and signal-to-noise ratio but requires strict usage conditions.
	
	In addition to addressing this practical problem, our objective is to propose a generalized algorithm for similar tasks that are crucial  but difficult to be accomplished directly, and explore how data from diverse modalities can facilitate in learning these hard tasks \citep{Li_Wei_Hong_Gong_2020, Afham_2022_CVPR, Ren_2021_CVPR}.
	
	We then proceed to investigate the underlying reasons behind the discriminative nature of features obtained from optical images in order to address this issue. As each dimension of the feature captures different semantics \citep{zeiler2014visualizing,zhou2016learning}, it is intuitively conjectured that the efficacy of optical features stems from their ability to capture a richer set of feature semantics, which is referred to as DS in this work. However, the abstract nature of DS requires further elaboration before it can be optimized.
	
	Regarding the output feature, we know that DS can be reflected in the correlation between the dimensions, where a lower channel-wise correlation implies a better DS. Furthermore, we consider the ability to express DS in the intermediate feature. It is widely acknowledged that the layer-by-layer encoding results of neural networks transition from generalized to task-specific, and more low-level visual information needs to be learned in the intermediate layer to facilitate the subsequent semantic extraction. In the meantime, the degree of scattering in spatial distribution is highly correlated with the underlying low-level visual information. Therefore, in the intermediate features, DS can be reflected in the spatial distribution of samples, i.e., the one with a closer instance-wise distance should have a better DS.
	
	Finally, we validate the aforementioned conjectures through motivating examples, which shows a robust correlation between feature discriminability and DS. Hence, we should aim to find the loss function that can 
	\begin{figure*}[!ht]
		\centering
		\includegraphics[width=0.95\textwidth]{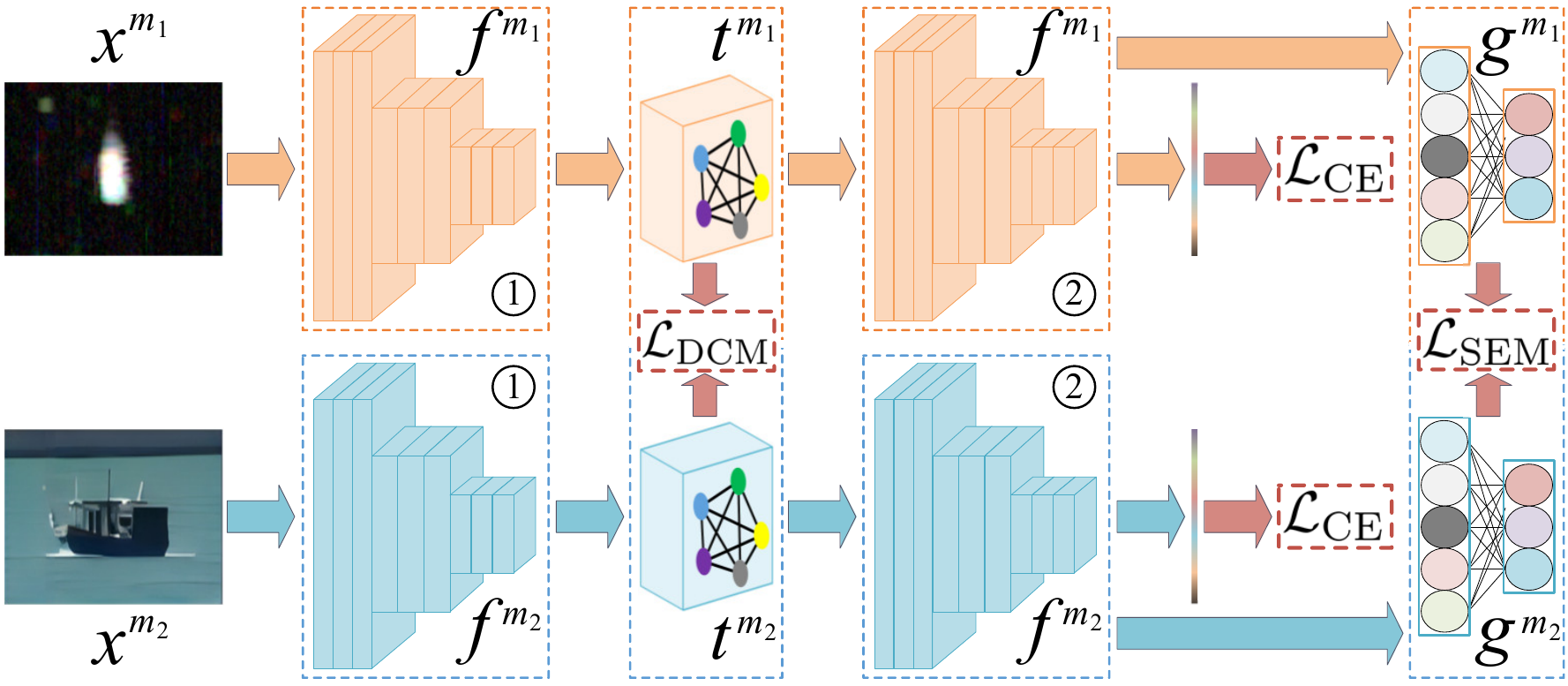}
		\caption{The overall pipeline of the proposed DS-based CMKD method. In this illustration, data streams from diverse modalities are delineated using distinct hues, \normalsize{\textcircled{\scriptsize{1}}}\normalsize\enspace and \normalsize{\textcircled{\scriptsize{2}}}\normalsize\enspace depict the initial and terminal phases of learning in encoders $f$. The loss function of the proposed method consists of three components, including $\mathcal{L}_{\text{CE}}$, $\mathcal{L}_{\text{SEM}}$ and $\mathcal{L}_{\text{DCM}}$. They align with the modules outlined in Subsection \ref{subsec:SRM} through \ref{subsec:DCM}, which are employed for receiving supervision, enhancing semantics, and calibrating distribution, respectively.}
		\label{fig:fram}
	\end{figure*}
	transfer the structural knowledge in DS to guarantee the effectiveness of CMKD methods in enhancing hard tasks. This guides us to propose the following approach.
	
	\section{Methodology}
	In this section, we present the formal definitions of channel-wise correlation and spatial distribution. Subsequently, we introduce our proposed DS-based CMKD method, which is comprised of three modules: the supervision reception module (SRM), the semantic enhancement module (SEM), and the distribution calibration module (DCM). The overall framework is illustrated in Fig. \ref{fig:fram}.
	
	\subsection{Supervision Reception Module}\label{subsec:SRM}
	This work falls into the study of supervised KD. So the SRM is employed to receive annotations from different tasks. We denote the input data as $X=\{(x_1^{m_1},x_1^{m_2},y_1),$\\$\cdots,(x_N^{m_1},x_N^{m_2},y_N)\}$, where $x_i^{m_1}$ and $x_i^{m_2}$ represent the $i$th paired cross-modal samples from the hard and simple tasks, and $y_i$ represents the label of the $i$th sample pair. To obtain corresponding feature representations, the input data is encoded into a latent space via the encoders $f^{m_1}$ and $f^{m_2}$. The following two classifiers, $h^{m_1}$ and $h^{m_2}$, are composed of fully connected layers. Based on the above, the loss function of SRM can be formulated as follows:
	\begin{equation} 
		\min_{f^{m_i},h^{m_i}} \ \mathcal{L}_{\text{CE}}(y, h^{m_i}(f^{m_i}(x^{m_i})))
		\label{srm}
	\end{equation} 
	where $\mathcal{L}_{\text{CE}}$ denotes the cross-entropy loss. Its specific definition is as follows:
	\begin{equation} 
		\mathcal{L}_{\text{CE}}(y, h(f(x))) = -\frac{1}{N}\sum_{i=1}^N\sum_{k=1}^K \mathbb I(y_i==k)\cdot \log h(f(x_i))
		\label{srm2}
	\end{equation}
	where $K$ denotes the total number of training categories, and $\mathbb I(\cdot)$ is the indicator function.
	
	\subsection{Semantic Enhancement Module}\label{subsec:SEM}
	In classical KD methods, the knowledge to be transferred is expressed as softened class scores, as follows:
	\begin{equation} 
		p(x_i,\tau)=\frac{\exp(h(f(x_i))/\tau)}{\sum_j\exp(h(f(x_j))/\tau)}
	\end{equation}
	where $\tau$ is a temperature factor to control the importance of each soft target. To enforce the logits-based knowledge transfer from the simple task to the hard one, the distillation loss is introduced as:
	\begin{equation} 
		\min_{f,h} \ \mathcal{L}_{\text{div}}(p(x^{m_1},\tau), p(x^{m_2},\tau))
		\label{sem_div}
	\end{equation}
	where $\mathcal{L}_{\text{div}}$ denotes the divergence measurement. Typically, Kullback-Leibler divergence, also known as relative entropy, has been commonly employed \citep{HintonKD}. Furthermore, the usage of mean square error (MSE) or earth mover's distance (EMD) to represent $\mathcal{L}_{\text{div}}$, and its application on the intermediate feature maps, have also been extensively explored in recent research works \citep{romero2014fitnets,10.1007/978-3-030-58595-2_40,chen2021crossSEMCKD}.
	
	The SEM is devised to transfer the DS-based structural knowledge obtained from output features, which can be seen as an alternative to Eq.  \eqref{sem_div}. Specifically, we employ two projectors $g^{m_1}$ and $g^{m_2}$ to model the dimensional information and map the encoded embeddings to higher dimensions \citep{chen2020simple}. The projected embeddings are represented as $z^{m_1}$ and $z^{m_2}$, where $z^{m_1}, z^{m_2}\in \mathbb{R}^{b\times d}$, $b$ and $d$ denote the number of samples in a mini-batch and feature maps in a projected embedding, respectively. A straightforward approach to measure the DS is to compute the divergence between their self-similarity matrices:
	\begin{equation} 
		\mathcal{L}_{\text{div}}(S(z^{m_1}), S(z^{m_2})) = \mathcal{L}_{\text{div}}((z^{m_1})^\top\cdot z^{m_1}, (z^{m_2})^\top\cdot z^{m_2}).
		\label{sem1}
	\end{equation}
	
	The above loss function necessitates the convergence of the two similarity matrices to equality at each position, which may not align with the expected objective of diminishing channel-wise correlation. To address this issue, a straightforward alternative is to employ the cross-correlation matrix $C$ of $z^{m_1}$ and $z^{m_2}$ in place of the self-similarity matrices. The cross-correlation matrix is defined as follows:
	\begin{equation} 
		\begin{split}
			C_{ij} & = \text{norm}({z^{m_1}_i})^\top\cdot \text{norm}({z^{m_2}_j})\\
			& = \frac{{\sum {_bz_{b,i}^{m_1} z_{b,j}^{m_2}} }}{{{{\sqrt {\sum\nolimits_b {(z_{b,i}^{m_1})^2} } }}{{\sqrt {\sum\nolimits_b {(z_{b,j}^{m_2})^2} } }}}}
		\end{split}
		\label{sem2}
	\end{equation}
	where $b$ is the index of min-batch and $i$, $j$ represent the row and column of $C$, also corresponding to the $i$th and $j$th dimension of $z^{m_1}$ and $z^{m_2}$. From the definition, it can be seen that the value of $C$ is between $-1$ to $1$, where $-1$ represents perfect anti-correlation, and $1$ represents perfect correlation. Based on Eq. \eqref{sem2} and let $\mathcal{L}_{\text{div}}$ to be MSE loss, Eq. \eqref{sem1} can be expressed as:
	\begin{equation} 
		\mathcal{L}_{\text{SEM}}(C) = \sum\limits_i {{{(1 - {C_{ii}})}^2}}  + \lambda \sum\limits_i {\sum\limits_{j \ne i} {C_{ij}^2} }
		\label{sem_revise}
	\end{equation}
	where $\lambda>0$ controls the trade-off. It can be observed from the definition that minimizing $\mathcal{L}_{\text{SEM}}$ requires the diagonal elements of $C$ to be equal to one, while simultaneously driving the off-diagonal elements of $C$ to zero. Taking $z_{b,i}^{m_1}$ as an example, Eq. \eqref{sem_revise} encourages $z_{b,i}^{m_1}$ to be as similar as possible to $z_{b,i}^{m_2}$, while being as dissimilar as possible to $z_{b,j}^{m_2}$ when $j\neq i$. The first term of the SEM loss ensures the cross-modal alignment of $z^{m_1}$ and $z^{m_2}$, while the second term allows $z^{m_1}$ to mimic the DS of $z^{m_2}$. Considering the low-level correlation between each dimension of $z^{m_2}$, the SEM explicitly enhances the variability of $z^{m_1}$, enabling it to learn more semantic information for better accuracy on the hard task.
	
	\subsection{Distribution Calibration Module}\label{subsec:DCM}
	In the previous subsection, we utilized the channel-wise correlation of the output feature as the carrier of DS to design the SEM. However, other concepts in intermediate layers can also impact the DS. After thorough analyses, we propose that the distribution of intermediate features is suitable for addressing this issue. Consequently, we construct the DCM to calibrate the distribution of intermediate features obtained from the hard task.
	
	Denote $t^{m_1}$ and $t^{m_2}$ as the intermediate features of $f^{m_1}$ and $f^{m_2}$, we want to find a functional metric that can be used to measure the distribution in the feature space. Inspired by \citet{wang2020understanding}, we define the metric as the logarithm of the average radial basis function (RBF), which can be written as:
	\begin{equation} 
		\mathcal{L}_{\text{DM}}(u,v) = \log\mathbb{E}_{u,v\sim p}[G_t(u,v)]
		\label{dcm_dis}
	\end{equation} 
	where $u,v\in \mathbb{R}^n$ and $p$ is the data distribution over $\mathbb{R}^n$, the logarithm operation is to ensure good computational properties, and $G_t$ is the RBF kernel, which can be defined as:
	\begin{equation} 
		G_t(u,v) = \exp(-\frac{\Vert t^{m_1}-t^{m_2} \Vert_2^2}{\sigma^2})
		\label{dcm_rbf}
	\end{equation} 
	where $\sigma$ is a fixed hyperparameter. Based on the above, we define the DCM loss as follows:
	\begin{equation} 
		\mathcal{L}_{\text{DCM}}(t^{m_1},t^{m_2}) = \mathcal{L}_{\text{DM}}(t^{m_1},t^{m_1}) + \mathcal{L}_{\text{DM}}(t^{m_1},t^{m_2}).
		\label{dcm_revise}
	\end{equation}
	
	The empirical risk of the terms in Eq. \eqref{dcm_revise} can be expressed as:
	\begin{equation}
		\mathcal{L}_{\text{DM}}(t^{m_q},t^{m_v}) = \log(\frac{1}{N^2}\sum_{i=1}^N\sum_{j=1}^N\exp(-\frac{\Vert t^{m_q}_i-t^{m_v}_j \Vert_2^2}{\sigma^2})).
		\label{dcm_revise2}
	\end{equation}
	
	This definition empirically computes the instance-wise distance between $t^{m_q}$ and $t^{m_v}$, which is expressed as an RBF kernel. As the number of instances $N$ tends towards infinity, minimizing Eq. \eqref{dcm_revise2} will lead the intermediate feature distribution $p_{\text{imd}}(t^{m_q},t^{m_v})$ to weakly converge to a uniform distribution \citep{wang2020understanding}. 
	
	From observing Eq. \eqref{dcm_revise}, it is evident that the former term optimizes the self-relative distance, while the latter term focuses on the cross-relative distance. Therefore, the first term of Eq. \eqref{dcm_revise} explicitly calibrates the distribution of $t^{m_1}$ by reducing self-relative instance-wise distances, leading to the distribution $p_{\text{imd}}(t^{m_1})$ approaching a uniform distribution. In contrast, the second term of Eq. \eqref{dcm_revise} shortens the distance between $t^{m_1}$ and $t^{m_2}$, aligning the distribution of  $p_{\text{imd}}(t^{m_1})$ and $p_{\text{imd}}(t^{m_2})$. The two items form a trade-off, ensuring that more low-level visual information can be retained while preventing $p_{\text{imd}}(t^{m_1})$ from deviating too much from $p_{\text{imd}}(t^{m_2})$. By combining the two terms, the DCM explicitly enhances the DS from an intermediate feature perspective, which consequently facilitates the learning of more semantics for the hard task.
	
	\subsection{Overall loss function}
	Utilizing the aforementioned SEM and DCM, the structural knowledge in DS can be conveyed through both output and intermediate perspectives. Specifically, the purpose of these modules is to decrease the interdependence among output channels and the distance among intermediate instances, with the goal of enhancing DS. Through the integration of richer semantic information via an improved DS, it is feasible to attain  improvements in performance for hard tasks.
	
	Given a mini-batch of training data, the loss function of the proposed DS-based CMKD method can be written as: 
	\begin{equation} \label{fun:overall}
		\min_{f^{m_1},h^{m_1},g^{m_1}} \ \mathcal{L}_{\text{ours}} = \mathcal{L}_{\text{CE}} + \gamma (\mathcal{L}_{\text{SEM}} + \mathcal{L}_{\text{DCM}})
	\end{equation}
	where $\gamma>0$ controls the ratio of two terms. Next, we analyze the relationship between the proposed approach and the relevant existing methods. 
	
	\textbf{Relation to logits-based KD.} \ Logits-based approaches aim to align cross-modal outputs by enforcing one to mimic the logits of the other \citep{HintonKD}. This idea can also be reflected in the proposed method, but the implementation is changed. Specifically, in the first term of $\mathcal{L}_{\text{SEM}}$, we utilize dot products as the similarity measurement to align cross-modal outputs.
	
	\textbf{Relation to feature-based KD.} \ In feature-based approaches \citep{romero2014fitnets,10.1007/978-3-030-58595-2_40}, numerical alignment of intermediate features is employed as a supplement to output mimicking. Our approach also aligns the intermediate features, as indicated in the second item of $\mathcal{L}_{\text{DCM}}$. However, this alignment is implemented based on spatial distribution and serves as a supplement to DS-based output mimicking.
	
	\textbf{Relation to adversarial-based KD.} Adversarial-based approaches incorporate a trainable discriminator to attain modal-invariant intermediate features \citep{GANKDwang2018adversarial} or logits \citep{xu2017training}. However, our objective is to attain cross-modal invariance of DS at both the intermediate and output aspects.
	
	\textbf{Relation to self-supervised-based KD.} The loss functions employed by self-supervised-based approaches \citep{xu2020knowledge} share certain similarities with our method. Nevertheless, the two are directed towards different forms of supervision.
	
	Lastly, although some other methods, e.g., relation-based \citep{park2019relational} and architecture-based \citep{8447210} methods, may not be directly linked to our proposed method, they can be readily incorporated in a plug-and-play manner.
	
	\begin{figure*}[!ht]
		\centering
		\includegraphics[width=0.95\textwidth]{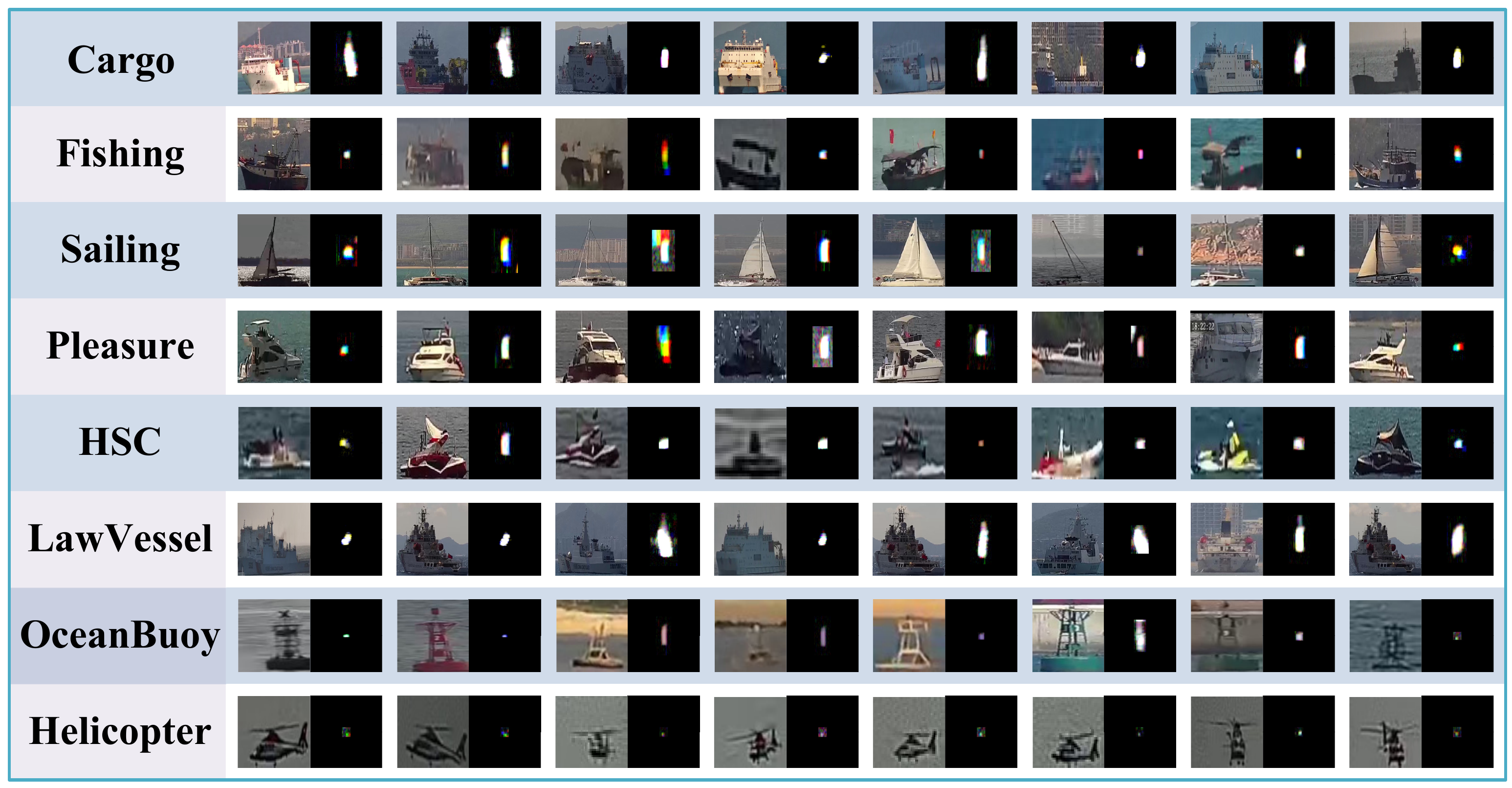}
		\caption{Examples of the collected real-world CML dataset consist of cargo, HSC, ocean buoy, helicopter, and four kinds of the vessel, including fishing vessel, sailing vessel, pleasure vessel, and law vessel.}
		\label{fig:data}
	\end{figure*}
	\begin{figure*}[!ht]
		\centering
		\includegraphics[width=0.9\textwidth]{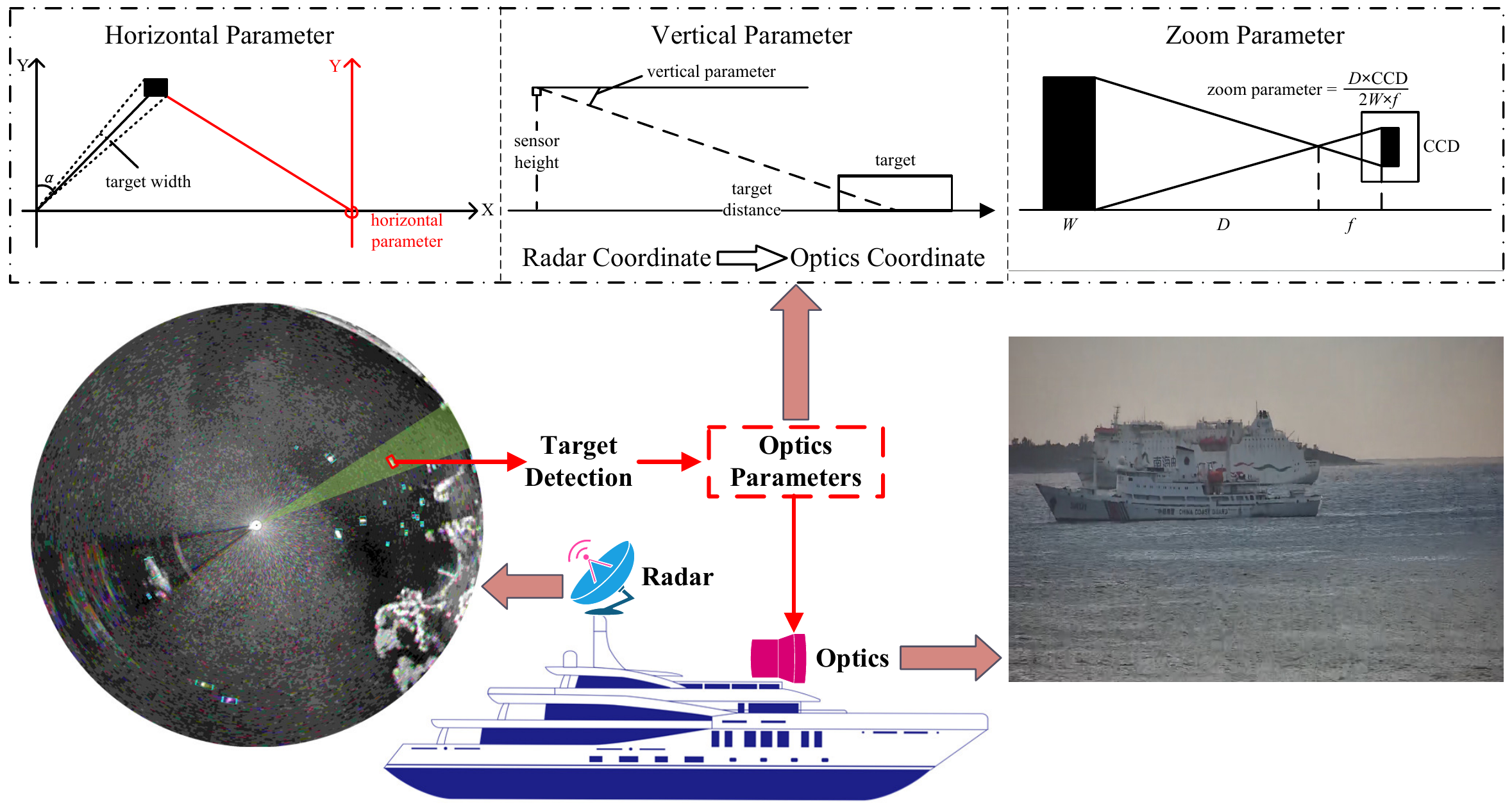}
		\caption{Illustration of steps for CML dataset collection, including radar target detection, optical parameter calculation, and paired cross-modal data acquisition. In this process, horizontal, vertical, and zoom parameters of the optics need to be calculated according to the longitude and latitude of the target detected by the radar and the relative relationship between the coordinates of radar and optics.}
		\label{fig:datacollect}
	\end{figure*}
	\section{Cross-Modal Learning Dataset}
	Paired cross-modal data collection and annotation are essential for developing CML algorithms. To address this issue, we establish an offshore workstation in Sanya, China, which equips with optical and radar sensors. A large number of sea trials have been conducted to collect sufficient data. Examples of the paired data are shown in Fig. \ref{fig:data}. This dataset is available at: \url{https://drive.google.com/file/d/1hBoLkS9uGKKzpsFCyhgr6jCgfY3I_eil/view?usp=share_link}.
	
	\subsection{Collection Procedure}
	We follow the subsequent steps to collect the paired optical and radar data, then construct the CML dataset.
	
	\textbf{Firstly}, radar information acquisition equipment can convert radar analog signals into radar echo intensity data at azimuth distance, and then generate radar B-scan images of consecutive frames. Then we use a YOLO-based target detection algorithm \citep{redmon2016you} to get the target position, longitude, latitude, and unique ID relative to the radar images.
	
	\textbf{Secondly}, the target position $(x, y, w, h)$ in the radar images can be used to calculate the azimuth distance $(A_1, D_1)$ of the target relative to the radar:
	\begin{equation}
		A_1=(x+\frac{w}{2})/X\times 360, \enspace D_1=(Y-y-\frac{h}{w})/Y\times R
	\end{equation}
	where the radar coordinate system is 0 degrees in the north direction, the scanning range is $360$ degrees, the radar range is $R$, and the radar image size is $X\times Y$. Through the target relative position $(A_1, D_1)$ and radar latitude and longitude $P_1$, the target latitude and longitude $P_2$ can be calculated, and then combined with the optical sensor latitude and longitude $P_3$, the target relative optical distance $(A_2, D_2)$ can be calculated.
	
	Through the relative position of the target $(A_2, D_2)$, we can calculate the optics parameters, including horizontal azimuth parameter $P$, vertical pitch parameter $T$, lens zoom multiplier parameter $Z$, to control the optical sensor to align with the radar target. The first two parameters can be calculated by:
	\begin{equation}
		P = A_2 + B, \enspace T = \arctan(L/D_2)
	\end{equation}
	where $B$ is the angle between the optics coordinate system 0 degrees and the north, and $L$ is the height of the optical device. The optics coordinate system is 0 degrees horizontal. The last parameter can be calculated by:
	\begin{equation}
		Z = D_2\times I/(2W\times f)
	\end{equation}
	where $I$ denotes the charge coupled device (CCD) length, $f$ represents the minimum focal length, the target imaging accounts for $1/2$ of the image. The actual length of the target is $W$ that can be deduced from the target width $w$ in radar images:
	\begin{equation}
		W = \arctan(180\frac{w}{X} -b)\times 2D_2
	\end{equation}
	where $b$ denotes radar beam width.
	
	\textbf{Thirdly}, we rotate the optical sensor and shoot the target continuously to generate the optical image of consecutive frames. Then another YOLO-based detector is used to get the target positions in optical modality.
	
	\textbf{Lastly}, we match the position of targets in radar and optics, and conclude the acquisition of paired cross-modal data.
	
	The whole paired data collection process is shown in Fig. \ref{fig:datacollect}. It shows the process of cross-modal target location matching and the calculation of each parameter in this process.
	
	\subsection{Dataset Characteristics}
	The radar image is collected by	Furuno FAR-2117. The pseudo-RGB image data is generated by splicing the original $4096\times 4096$ radar B-scan echo. Three consecutive frames constitute the RGB channels of a radar image. By superimposing more consecutive frames of radar echo images, radar N-frame time series images can be formed. The optical image is collected from a Hikvision camera, and the specific type is DS-2TD9167-230ZG2F. 
	
	The collected real-world CML dataset contains 8 categories, including cargo, fishing vessel, sailing vessel, pleasure vessel, high-speed catamaran (HSC), law vessel, ocean buoy, and helicopter. The number of each category is shown in Table \ref{label:number}. 
	\begin{table}[!ht]
		\centering
		\caption{The proportion of different categories in the collected real-world CML dataset}
		\resizebox{0.9\columnwidth}{!}
		{
			\begin{tabular}{lcc}
				
				\toprule
				Category & Number& Percentage\\
				\midrule
				Cargo           & 9194   &34.8\% \\
				
				Fishing Vessel   &4054  &15.3\%\\
				
				Sailing Vessel        & 1564   & 6.1\%\\
				
				Pleasure Vessel        & 5964  &22.6 \%\\
				
				HSC            &3228    & 12.2\% \\
				Law Vessel      &1084    &  4.1\% \\
				
				Ocean Buoy      & 654   &2.5 \%\\
				Helicopter       &630    &2.4 \%\\
				\bottomrule
			\end{tabular}
		}
		\label{label:number}
	\end{table}
	
	\section{Experiments}\label{sec:exp}
	This section aims to evaluate the performance of the proposed method using a real-world CML dataset and four benchmark datasets. To achieve this goal, we compare our approach with several state-of-the-art KD methods chosen in the experiment.
	
	\subsection{Datasets}
	We use seven datasets in our experiments, including one real-world CML dataset proposed in this paper, and six benchmark datasets. Their specific information is as follows:
	
	\textbf{Radar and optics CML dataset} is collected by us that comprises 8 categories, where each category consists of a pair of data modalities that pertain to the same target: radar and optics. Notably, the size of each paired data is 224$\times$224.
	
	\textbf{CIFAR-100} is another prevalent benchmark dataset for image classification tasks. It includes 100 categories, and each category contains 5,000 images, with the size of each image being 32$\times$32.
	
	\textbf{CIFAR-10} is a well-established benchmark dataset that is extensively employed in image classification tasks. It encompasses ten categories, with each category containing 5,000 images. The size of each image in the dataset is 32$\times$32.
	
	\textbf{STL-10} is collected inspired by the CIFAR-10 dataset, with additional modifications and improvements. In comparison to the previously mentioned datasets, it has a higher resolution and can be used to develop more scalable deep learning-based methods.
	
	\textbf{Tiny ImageNet} is an image classification dataset that comprises 200 distinct classes, each of which has 500 training images, 50 validation images, and 50 test images. It is a subset of ImageNet and is released by Stanford University in 2016.
	
	\textbf{ImageNet-100} is a selection of 100 categories from the ImageNet dataset, each containing 1,000 images with the size of 224$\times$224.
	
	\textbf{ImageNet} is collected from various sources on the internet, and each image is labeled with a category label from the WordNet hierarchy. It is a widely recognized large-scale dataset that includes more than 21,000 categories, comprising 1.3 million training images and 50,000 test images.
	
	The collected CML dataset comprises both radar and optics modalities, wherein the former has relatively lower quality compared to the latter. The direct use of the dataset is suitable for verifying the performance of CML methods. However, in the six benchmark datasets, only one modality of optics is incorporated. To emulate the practical application scenario of CML methods, we introduce noise to the original data by adding a pretzel noise, thereby forming two sets of data, i.e., high and low quality. The signal-to-noise ratio for adding pretzel noise to the first four datasets is set to 0.1, while the signal-to-noise ratio on ImageNet-100 and ImageNet datasets is set to 0.5.	
	
	\subsection{Experimental Settings}
	The experiments follow the standard KD pipeline: Aiming at modalities with high data quality, we train a teacher network. Following this, we freeze the parameters of the teacher network and proceed to train a student network that is specifically designed for modalities with lower data quality, while being guided by the teacher network. We choose ResNet32$\times$4 and VGG13 as the backbone for teacher, and ResNet8$\times$4, ShuffleNet, VGG8 and MobileNet as the backbone for student. Stochastic gradient descent (SGD) with the momentum of 0.9 is used to minimize the objective functions. During training, the involved models are trained for 300 epochs with an initial learning rate of 0.1. The weight decay is set as 5e-4. The learning rate decreases to 1e-6 at the end of training. PyTorch is chosen as the primary software tool for conducting our experiments, while four RTX TITAN GPUs are employed for computation from a hardware standpoint.
	
	We compare our proposed method with the state-of-the-art methods, which include logit-based, feature-based, and relation-based KD methods. The first category consists of vanilla KD (KD) \citep{HintonKD} and adversarial logits-based KD (ALKD) \citep{xu2017training}. The second category consists of Fitnets \citep{romero2014fitnets}, attention transfer (AT) \citep{ATkomodakis2017paying}, adversarial feature-based KD (AFKD) \citep{GANKDwang2018adversarial}, neuron selectivity transfer (NST) \citep{Huang2017arxiv}, variational information distillation (VID) \citep{VIDahn2019variational}, contrastive representation distillation (CRD) \citep{Tian2020CRD}, semantic calibration based KD (SemCKD) \citep{chen2021crossSEMCKD} in the second one. The third category consists of relational KD (RKD) \citep{park2019relational}, flow of solution procedure matrix based KD (FSP) \citep{8100237FSP} and correlation congruence based KD (CC) \citep{CCpeng2019correlation}.
	
	\subsection{Experimental Results}
	Tables \ref{table:oursdata}-\ref{table:Imagenet} show the top-1 accuracy of CML on seven datasets for all the methods involved in the experiments. Extensive experimental results verify the effectiveness of the proposed method. Below we give the specific analysis of the experimental results on each dataset.
	\begin{table*}[!ht]  
		\centering
		\caption{The classification top-1 accuracy results on the real-world CML dataset with ResNet32$\times$4 and VGG13 as the teacher backbone, and with ResNet8$\times$4, ShuffleNetV1, ShuffleNetV2, VGG8 and MobileNetV2 as the student backbone.}
		\resizebox{0.8\textwidth}{!}
		{
			\begin{tabular}{lcccccccccc}
				\toprule
				Teacher& \multicolumn{2}{c}{ResNet32$\times$4} &  \multicolumn{2}{c}{ResNet32$\times$4}&\multicolumn{2}{c}{ResNet32$\times$4}&\multicolumn{2}{c}{VGG13} & \multicolumn{2}{c}{VGG13} \\
				Student& \multicolumn{2}{c}{ResNet8$\times$4} &  \multicolumn{2}{c}{ShuffleNetV1}&\multicolumn{2}{c}{ShuffleNetV2}&\multicolumn{2}{c}{VGG8} & \multicolumn{2}{c}{MobileNetV2} \\
				\midrule
				Baseline-T&\multicolumn{2}{c}{99.23}&\multicolumn{2}{c}{99.23}&\multicolumn{2}{c}{99.23}&\multicolumn{2}{c}{97.02}&\multicolumn{2}{c}{97.02}\\
				Baseline-S&\multicolumn{2}{c}{84.00}&\multicolumn{2}{c}{80.31}&\multicolumn{2}{c}{79.43}&\multicolumn{2}{c}{80.01}&\multicolumn{2}{c}{77.82}\\
				\midrule
				KD&\multicolumn{2}{c}{88.29($\uparrow$4.29)}&\multicolumn{2}{c}{86.45($\uparrow$6.14)}&\multicolumn{2}{c}{84.77($\uparrow$5.34)}&\multicolumn{2}{c}{82.89($\uparrow$2.88)}&\multicolumn{2}{c}{80.16($\uparrow$2.34)}\\
				ALKD&\multicolumn{2}{c}{87.15($\uparrow$3.15)}&\multicolumn{2}{c}{86.16($\uparrow$5.85)}&\multicolumn{2}{c}{84.56($\uparrow$5.13)}&\multicolumn{2}{c}{82.13($\uparrow$2.12)}&\multicolumn{2}{c}{79.89($\uparrow$2.07)}\\
				Fitnets&\multicolumn{2}{c}{87.89($\uparrow$3.89)}&\multicolumn{2}{c}{86.89($\uparrow$6.58)}&\multicolumn{2}{c}{83.46($\uparrow$4.03)}&\multicolumn{2}{c}{82.49($\uparrow$2.48)}&\multicolumn{2}{c}{78.49($\uparrow$0.67)}\\
				AT&\multicolumn{2}{c}{87.53($\uparrow$3.53)}&\multicolumn{2}{c}{87.16($\uparrow$6.85)}&\multicolumn{2}{c}{84.79($\uparrow$5.36)}&\multicolumn{2}{c}{82.46($\uparrow$2.45)}&\multicolumn{2}{c}{79.73($\uparrow$1.91)}\\
				AFKD&\multicolumn{2}{c}{87.99($\uparrow$3.99)}&\multicolumn{2}{c}{86.65($\uparrow$6.34)}&\multicolumn{2}{c}{84.92($\uparrow$5.49)}&\multicolumn{2}{c}{81.49($\uparrow$1.48)}&\multicolumn{2}{c}{80.06($\uparrow$2.24)}\\
				NST&\multicolumn{2}{c}{88.93($\uparrow$4.93)}&\multicolumn{2}{c}{87.93($\uparrow$7.62)}&\multicolumn{2}{c}{85.46($\uparrow$6.03)}&\multicolumn{2}{c}{83.49($\uparrow$3.48)}&\multicolumn{2}{c}{80.89($\uparrow$3.07)}\\
				VID&\multicolumn{2}{c}{88.36($\uparrow$4.36)}&\multicolumn{2}{c}{87.56($\uparrow$7.25)}&\multicolumn{2}{c}{85.65($\uparrow$6.22)}&\multicolumn{2}{c}{83.79($\uparrow$3.78)}&\multicolumn{2}{c}{80.76($\uparrow$2.94)}\\
				CRD&\multicolumn{2}{c}{89.48($\uparrow$5.48)}&\multicolumn{2}{c}{87.16($\uparrow$6.85)}&\multicolumn{2}{c}{86.15($\uparrow$6.72)}&\multicolumn{2}{c}{83.15($\uparrow$3.14)}&\multicolumn{2}{c}{81.01($\uparrow$3.19)}\\
				SemCKD&\multicolumn{2}{c}{89.13($\uparrow$5.13)}&\multicolumn{2}{c}{87.89($\uparrow$7.58)}&\multicolumn{2}{c}{86.23($\uparrow$6.80)}&\multicolumn{2}{c}{83.58($\uparrow$3.57)}&\multicolumn{2}{c}{81.16($\uparrow$3.34)}\\
				RKD&\multicolumn{2}{c}{88.16($\uparrow$4.16)}&\multicolumn{2}{c}{86.46($\uparrow$6.15)}&\multicolumn{2}{c}{85.98($\uparrow$6.55)}&\multicolumn{2}{c}{82.98($\uparrow$2.97)}&\multicolumn{2}{c}{80.67($\uparrow$2.85)}\\
				FSP&\multicolumn{2}{c}{88.66($\uparrow$4.66)}&\multicolumn{2}{c}{86.71($\uparrow$6.40)}&\multicolumn{2}{c}{85.42($\uparrow$5.99)}&\multicolumn{2}{c}{81.79($\uparrow$1.78)}&\multicolumn{2}{c}{79.83($\uparrow$2.01)}\\
				CC&\multicolumn{2}{c}{88.62($\uparrow$4.62)}&\multicolumn{2}{c}{86.03($\uparrow$5.72)}&\multicolumn{2}{c}{85.78($\uparrow$6.35)}&\multicolumn{2}{c}{82.95($\uparrow$2.94)}&\multicolumn{2}{c}{80.22($\uparrow$2.40)}\\
				\midrule
				Ours&\multicolumn{2}{c}{\textbf{89.87}($\uparrow$\textbf{5.87})}&\multicolumn{2}{c}{\textbf{88.02}($\uparrow$\textbf{7.71})}&\multicolumn{2}{c}{\textbf{86.94}($\uparrow$\textbf{7.51})}&\multicolumn{2}{c}{\textbf{84.01}($\uparrow$\textbf{4.00})}&\multicolumn{2}{c}{\textbf{81.88}($\uparrow$\textbf{4.06})}\\
				\bottomrule
			\end{tabular}
		}
		\label{table:oursdata}
	\end{table*}
	\begin{table*}[!t]  
		\centering
		\caption{The classification top-1 accuracy results on CIFAR-100 and CIFAR-10 datasets with ResNet32$\times$4 as the teacher backbone, and with ResNet8$\times$4, ShuffleNetV1, ShuffleNetV2 as the student backbone.}
		\resizebox{0.96\textwidth}{!}
		{
			\begin{tabular}{lccccccccccccc}
				\toprule
				& \multicolumn{6}{c}{CIFAR-100} && \multicolumn{6}{c}{CIFAR-10} \\
				\cline{2-7} \cline{9-14}
				
				Teacher& \multicolumn{2}{c}{ResNet32$\times$4} &  \multicolumn{2}{c}{ResNet32$\times$4}&\multicolumn{2}{c}{ResNet32$\times$4}&&\multicolumn{2}{c}{ResNet32$\times$4} & \multicolumn{2}{c}{ResNet32$\times$4}&\multicolumn{2}{c}{ResNet32$\times$4} \\
				Student& \multicolumn{2}{c}{ResNet8$\times$4} &  \multicolumn{2}{c}{ShuffleNetV1}&\multicolumn{2}{c}{ShuffleNetV2}&&\multicolumn{2}{c}{ResNet8$\times$4} & \multicolumn{2}{c}{ShuffleNetV1}&\multicolumn{2}{c}{ShuffleNetV2} \\
				\midrule
				Baseline-T&\multicolumn{2}{c}{79.42}&\multicolumn{2}{c}{79.42}&\multicolumn{2}{c}{79.42}&&\multicolumn{2}{c}{93.60}&\multicolumn{2}{c}{93.60}&\multicolumn{2}{c}{93.60}\\
				Baseline-S&\multicolumn{2}{c}{72.49}&\multicolumn{2}{c}{70.51}&\multicolumn{2}{c}{71.82}&&\multicolumn{2}{c}{81.40}&\multicolumn{2}{c}{79.81}&\multicolumn{2}{c}{80.32}\\
				\midrule
				KD&\multicolumn{2}{c}{73.66($\uparrow$1.17)}&\multicolumn{2}{c}{71.68($\uparrow$1.17)}&\multicolumn{2}{c}{74.33($\uparrow$2.51)}&&\multicolumn{2}{c}{83.50($\uparrow$2.10)}&\multicolumn{2}{c}{83.42($\uparrow$3.61)}&\multicolumn{2}{c}{82.99($\uparrow$2.67)}\\
				ALKD&\multicolumn{2}{c}{72.92($\uparrow$0.43)}&\multicolumn{2}{c}{71.03($\uparrow$0.52)}&\multicolumn{2}{c}{73.32($\uparrow$1.50)}&&\multicolumn{2}{c}{82.33($\uparrow$0.93)}&\multicolumn{2}{c}{83.12($\uparrow$3.31)}&\multicolumn{2}{c}{82.78($\uparrow$2.46)}\\
				Fitnets&\multicolumn{2}{c}{73.72($\uparrow$1.23)}&\multicolumn{2}{c}{71.79($\uparrow$1.28)}&\multicolumn{2}{c}{74.23($\uparrow$2.41)}&&\multicolumn{2}{c}{83.52($\uparrow$2.12)}&\multicolumn{2}{c}{83.49($\uparrow$3.68)}&\multicolumn{2}{c}{82.89($\uparrow$2.57)}\\
				AT&\multicolumn{2}{c}{73.58($\uparrow$1.09)}&\multicolumn{2}{c}{71.27($\uparrow$0.76)}&\multicolumn{2}{c}{74.32($\uparrow$2.50)}&&\multicolumn{2}{c}{83.71($\uparrow$2.31)}&\multicolumn{2}{c}{83.89($\uparrow$4.08)}&\multicolumn{2}{c}{82.49($\uparrow$2.17)}\\
				AFKD&\multicolumn{2}{c}{73.95($\uparrow$1.46)}&\multicolumn{2}{c}{70.96($\uparrow$0.45)}&\multicolumn{2}{c}{73.19($\uparrow$1.37)}&&\multicolumn{2}{c}{83.14($\uparrow$1.74)}&\multicolumn{2}{c}{83.78($\uparrow$3.97)}&\multicolumn{2}{c}{83.33($\uparrow$3.01)}\\
				NST&\multicolumn{2}{c}{74.17($\uparrow$1.68)}&\multicolumn{2}{c}{71.99($\uparrow$1.48)}&\multicolumn{2}{c}{74.81($\uparrow$2.99)}&&\multicolumn{2}{c}{83.79($\uparrow$2.39)}&\multicolumn{2}{c}{84.12($\uparrow$4.31)}&\multicolumn{2}{c}{83.12($\uparrow$2.80)}\\
				VID&\multicolumn{2}{c}{74.24($\uparrow$1.75)}&\multicolumn{2}{c}{71.93($\uparrow$1.42)}&\multicolumn{2}{c}{73.12($\uparrow$1.30)}&&\multicolumn{2}{c}{83.82($\uparrow$2.42)}&\multicolumn{2}{c}{84.95($\uparrow$5.14)}&\multicolumn{2}{c}{83.06($\uparrow$2.74)}\\
				CRD&\multicolumn{2}{c}{74.51($\uparrow$2.02)}&\multicolumn{2}{c}{72.01($\uparrow$1.50)}&\multicolumn{2}{c}{75.33($\uparrow$3.51)}&&\multicolumn{2}{c}{84.01($\uparrow$2.61)}&\multicolumn{2}{c}{84.25($\uparrow$4.44)}&\multicolumn{2}{c}{83.49($\uparrow$3.17)}\\
				SemCKD&\multicolumn{2}{c}{74.73($\uparrow$2.24)}&\multicolumn{2}{c}{72.17($\uparrow$1.66)}&\multicolumn{2}{c}{75.85($\uparrow$4.03)}&&\multicolumn{2}{c}{84.20($\uparrow$2.80)}&\multicolumn{2}{c}{84.03($\uparrow$4.22)}&\multicolumn{2}{c}{83.26($\uparrow$2.94)}\\
				RKD&\multicolumn{2}{c}{73.98($\uparrow$1.49)}&\multicolumn{2}{c}{71.35($\uparrow$0.84)}&\multicolumn{2}{c}{74.96($\uparrow$3.14)}&&\multicolumn{2}{c}{82.96($\uparrow$1.56)}&\multicolumn{2}{c}{84.15($\uparrow$4.34)}&\multicolumn{2}{c}{83.98($\uparrow$3.66)}\\
				FSP&\multicolumn{2}{c}{73.56($\uparrow$1.07)}&\multicolumn{2}{c}{71.46($\uparrow$0.95)}&\multicolumn{2}{c}{74.14($\uparrow$2.32)}&&\multicolumn{2}{c}{83.17($\uparrow$1.77)}&\multicolumn{2}{c}{83.78($\uparrow$3.97)}&\multicolumn{2}{c}{83.23($\uparrow$2.91)}\\
				CC&\multicolumn{2}{c}{74.10($\uparrow$1.61)}&\multicolumn{2}{c}{71.88($\uparrow$1.37)}&\multicolumn{2}{c}{75.01($\uparrow$3.19)}&&\multicolumn{2}{c}{83.44($\uparrow$2.04)}&\multicolumn{2}{c}{83.47($\uparrow$3.66)}&\multicolumn{2}{c}{83.77($\uparrow$3.45)}\\
				\midrule
				Ours&\multicolumn{2}{c}{\textbf{75.11}($\uparrow$\textbf{2.62})}&\multicolumn{2}{c}{\textbf{72.86}($\uparrow$\textbf{2.35})}&\multicolumn{2}{c}{\textbf{76.02}($\uparrow$\textbf{4.20})}&&\multicolumn{2}{c}{\textbf{84.70}($\uparrow$\textbf{3.30})}&\multicolumn{2}{c}{\textbf{85.24}($\uparrow$\textbf{5.43})}&\multicolumn{2}{c}{\textbf{84.56}($\uparrow$\textbf{4.24})}\\
				\bottomrule
			\end{tabular}
		}
		\label{table:CIFAR}
	\end{table*}
	\begin{table*}[!h]  
		\centering
		\caption{The classification top-1 accuracy results on STL-10 and Tiny-ImageNet datasets with ResNet32$\times$4 as the teacher backbone, and with ResNet8$\times$4, ShuffleNetV1, ShuffleNetV2 as the student backbone.}
		\resizebox{0.96\textwidth}{!}
		{
			\begin{tabular}{lccccccccccccc}
				\toprule
				& \multicolumn{6}{c}{STL-10} && \multicolumn{6}{c}{Tiny-ImageNet} \\
				\cline{2-7} \cline{9-14}
				
				Teacher& \multicolumn{2}{c}{ResNet32$\times$4} &  \multicolumn{2}{c}{ResNet32$\times$4}&\multicolumn{2}{c}{ResNet32$\times$4}&&\multicolumn{2}{c}{ResNet32$\times$4} & \multicolumn{2}{c}{ResNet32$\times$4}&\multicolumn{2}{c}{ResNet32$\times$4} \\
				Student& \multicolumn{2}{c}{ResNet8$\times$4} &  \multicolumn{2}{c}{ShuffleNetV1}&\multicolumn{2}{c}{ShuffleNetV2}&&\multicolumn{2}{c}{ResNet8$\times$4} & \multicolumn{2}{c}{ShuffleNetV1}&\multicolumn{2}{c}{ShuffleNetV2} \\
				\midrule
				Baseline-T&\multicolumn{2}{c}{61.29}&\multicolumn{2}{c}{61.29}&\multicolumn{2}{c}{61.29}&&\multicolumn{2}{c}{76.43}&\multicolumn{2}{c}{76.43}&\multicolumn{2}{c}{76.43}\\
				Baseline-S&\multicolumn{2}{c}{52.93}&\multicolumn{2}{c}{49.88}&\multicolumn{2}{c}{50.25}&&\multicolumn{2}{c}{60.58}&\multicolumn{2}{c}{58.11}&\multicolumn{2}{c}{59.39}\\
				\midrule
				KD&\multicolumn{2}{c}{54.44($\uparrow$1.51)}&\multicolumn{2}{c}{54.92($\uparrow$5.04)}&\multicolumn{2}{c}{55.03($\uparrow$4.78)}&&\multicolumn{2}{c}{61.35($\uparrow$0.77)}&\multicolumn{2}{c}{62.03($\uparrow$3.92)}&\multicolumn{2}{c}{62.33($\uparrow$2.94)}\\
				ALKD&\multicolumn{2}{c}{53.19($\uparrow$0.26)}&\multicolumn{2}{c}{54.02($\uparrow$4.14)}&\multicolumn{2}{c}{54.02($\uparrow$3.77)}&&\multicolumn{2}{c}{61.89($\uparrow$1.31)}&\multicolumn{2}{c}{61.79($\uparrow$3.68)}&\multicolumn{2}{c}{61.89($\uparrow$2.50)}\\
				Fitnets&\multicolumn{2}{c}{53.86($\uparrow$0.93)}&\multicolumn{2}{c}{54.59($\uparrow$4.71)}&\multicolumn{2}{c}{54.89($\uparrow$4.64)}&&\multicolumn{2}{c}{61.78($\uparrow$1.20)}&\multicolumn{2}{c}{61.99($\uparrow$3.88)}&\multicolumn{2}{c}{62.85($\uparrow$3.46)}\\
				AT&\multicolumn{2}{c}{54.27($\uparrow$1.34)}&\multicolumn{2}{c}{54.61($\uparrow$4.73)}&\multicolumn{2}{c}{54.49($\uparrow$4.24)}&&\multicolumn{2}{c}{62.47($\uparrow$1.89)}&\multicolumn{2}{c}{62.89($\uparrow$4.78)}&\multicolumn{2}{c}{62.15($\uparrow$2.76)}\\
				AFKD&\multicolumn{2}{c}{53.15($\uparrow$0.22)}&\multicolumn{2}{c}{54.95($\uparrow$5.07)}&\multicolumn{2}{c}{55.02($\uparrow$4.77)}&&\multicolumn{2}{c}{62.45($\uparrow$1.87)}&\multicolumn{2}{c}{62.88($\uparrow$4.77)}&\multicolumn{2}{c}{62.89($\uparrow$3.50)}\\
				NST&\multicolumn{2}{c}{54.51($\uparrow$1.58)}&\multicolumn{2}{c}{54.49($\uparrow$4.61)}&\multicolumn{2}{c}{55.21($\uparrow$4.96)}&&\multicolumn{2}{c}{62.49($\uparrow$1.91)}&\multicolumn{2}{c}{63.16($\uparrow$5.05)}&\multicolumn{2}{c}{64.19($\uparrow$4.80)}\\
				VID&\multicolumn{2}{c}{54.37($\uparrow$1.44)}&\multicolumn{2}{c}{54.01($\uparrow$4.13)}&\multicolumn{2}{c}{55.41($\uparrow$5.16)}&&\multicolumn{2}{c}{61.49($\uparrow$0.91)}&\multicolumn{2}{c}{62.89($\uparrow$4.78)}&\multicolumn{2}{c}{64.90($\uparrow$5.51)}\\
				CRD&\multicolumn{2}{c}{54.62($\uparrow$1.69)}&\multicolumn{2}{c}{54.89($\uparrow$5.01)}&\multicolumn{2}{c}{55.46($\uparrow$5.21)}&&\multicolumn{2}{c}{62.77($\uparrow$2.19)}&\multicolumn{2}{c}{63.49($\uparrow$5.38)}&\multicolumn{2}{c}{63.14($\uparrow$3.75)}\\
				SemCKD&\multicolumn{2}{c}{54.78($\uparrow$1.85)}&\multicolumn{2}{c}{55.30($\uparrow$5.42)}&\multicolumn{2}{c}{54.95($\uparrow$4.70)}&&\multicolumn{2}{c}{62.93($\uparrow$2.35)}&\multicolumn{2}{c}{63.78($\uparrow$5.67)}&\multicolumn{2}{c}{64.12($\uparrow$4.73)}\\
				RKD&\multicolumn{2}{c}{53.82($\uparrow$0.89)}&\multicolumn{2}{c}{55.06($\uparrow$5.18)}&\multicolumn{2}{c}{54.48($\uparrow$4.23)}&&\multicolumn{2}{c}{61.45($\uparrow$0.87)}&\multicolumn{2}{c}{62.49($\uparrow$4.38)}&\multicolumn{2}{c}{63.49($\uparrow$4.10)}\\
				FSP&\multicolumn{2}{c}{54.15($\uparrow$1.22)}&\multicolumn{2}{c}{54.13($\uparrow$4.25)}&\multicolumn{2}{c}{54.89($\uparrow$4.64)}&&\multicolumn{2}{c}{61.84($\uparrow$1.26)}&\multicolumn{2}{c}{62.78($\uparrow$4.67)}&\multicolumn{2}{c}{63.87($\uparrow$4.48)}\\
				CC&\multicolumn{2}{c}{54.33($\uparrow$1.40)}&\multicolumn{2}{c}{53.49($\uparrow$3.61)}&\multicolumn{2}{c}{54.56($\uparrow$4.31)}&&\multicolumn{2}{c}{62.23($\uparrow$1.65)}&\multicolumn{2}{c}{61.48($\uparrow$3.37)}&\multicolumn{2}{c}{62.79($\uparrow$3.40)}\\
				\midrule
				Ours&\multicolumn{2}{c}{\textbf{55.02}($\uparrow$\textbf{2.09})}&\multicolumn{2}{c}{\textbf{55.66}($\uparrow$\textbf{5.78})}&\multicolumn{2}{c}{\textbf{55.92}($\uparrow$\textbf{5.67})}&&\multicolumn{2}{c}{\textbf{63.41}($\uparrow$\textbf{2.83})}&\multicolumn{2}{c}{\textbf{64.14}($\uparrow$\textbf{6.03})}&\multicolumn{2}{c}{\textbf{65.66}($\uparrow$\textbf{6.27})}\\
				\bottomrule
			\end{tabular}
		}
		\label{table:StlImagenet}
	\end{table*}
	
	\textbf{Results on the proposed CML dataset.} The experimental results on our collected real-world CML dataset are presented in Table \ref{table:oursdata}. The results demonstrate that the incorporation of CMKD techniques in different network architectures leads to improved accuracy in radar target recognition. Notably, the proposed method demonstrates superior performance improvement compared to other methods. Specifically, when ResNet32$\times$4 and ShuffleNetV1 are used as the backbone for optical and radar recognition, respectively, the accuracy of improvement reaches $7.71\%$, which is almost $10\%$ of the initial accuracy, thereby affirming the effectiveness of the proposed DS-based structural knowledge transfer method.
	\begin{table*}[!ht]  
		\centering
		\caption{The classification top-1 accuracy results on ImageNet-100 and ImageNet datasets with ResNet32$\times$4 as the teacher backbone, and with ResNet8$\times$4, ShuffleNetV1, ShuffleNetV2 as the student backbone.}
		\resizebox{0.96\textwidth}{!}
		{
			\begin{tabular}{lccccccccccccc}
				\toprule
				& \multicolumn{6}{c}{ImageNet-100} && \multicolumn{6}{c}{ImageNet} \\
				\cline{2-7} \cline{9-14}
				
				Teacher& \multicolumn{2}{c}{ResNet32$\times$4} &  \multicolumn{2}{c}{ResNet32$\times$4}&\multicolumn{2}{c}{ResNet32$\times$4}&&\multicolumn{2}{c}{ResNet32$\times$4} & \multicolumn{2}{c}{ResNet32$\times$4}&\multicolumn{2}{c}{ResNet32$\times$4} \\
				Student& \multicolumn{2}{c}{ResNet8$\times$4} &  \multicolumn{2}{c}{ShuffleNetV1}&\multicolumn{2}{c}{ShuffleNetV2}&&\multicolumn{2}{c}{ResNet8$\times$4} & \multicolumn{2}{c}{ShuffleNetV1}&\multicolumn{2}{c}{ShuffleNetV2} \\
				\midrule
				Baseline-T&\multicolumn{2}{c}{83.31}&\multicolumn{2}{c}{83.31}&\multicolumn{2}{c}{83.31}&&\multicolumn{2}{c}{76.28}&\multicolumn{2}{c}{76.28}&\multicolumn{2}{c}{76.28}\\
				Baseline-S&\multicolumn{2}{c}{64.26}&\multicolumn{2}{c}{61.86}&\multicolumn{2}{c}{62.67}&&\multicolumn{2}{c}{58.69}&\multicolumn{2}{c}{56.71}&\multicolumn{2}{c}{56.31}\\
				\midrule
				KD&\multicolumn{2}{c}{67.38($\uparrow$3.12)}&\multicolumn{2}{c}{66.10($\uparrow$4.24)}&\multicolumn{2}{c}{66.22($\uparrow$3.55)}&&\multicolumn{2}{c}{61.45($\uparrow$2.76)}&\multicolumn{2}{c}{61.11($\uparrow$4.40)}&\multicolumn{2}{c}{61.07($\uparrow$4.76)}\\
				ALKD&\multicolumn{2}{c}{66.58($\uparrow$2.32)}&\multicolumn{2}{c}{65.43($\uparrow$3.57)}&\multicolumn{2}{c}{66.01($\uparrow$3.34)}&&\multicolumn{2}{c}{60.59($\uparrow$1.90)}&\multicolumn{2}{c}{61.34($\uparrow$4.63)}&\multicolumn{2}{c}{61.01($\uparrow$4.70)}\\
				Fitnets&\multicolumn{2}{c}{66.92($\uparrow$2.66)}&\multicolumn{2}{c}{65.27($\uparrow$3.42)}&\multicolumn{2}{c}{65.87($\uparrow$3.20)}&&\multicolumn{2}{c}{60.92($\uparrow$2.23)}&\multicolumn{2}{c}{61.05($\uparrow$4.34)}&\multicolumn{2}{c}{61.23($\uparrow$4.92)}\\
				AT&\multicolumn{2}{c}{67.29($\uparrow$3.03)}&\multicolumn{2}{c}{65.84($\uparrow$3.98)}&\multicolumn{2}{c}{65.93($\uparrow$3.26)}&&\multicolumn{2}{c}{60.89($\uparrow$2.20)}&\multicolumn{2}{c}{61.41($\uparrow$4.70)}&\multicolumn{2}{c}{61.18($\uparrow$4.87)}\\
				AFKD&\multicolumn{2}{c}{67.01($\uparrow$2.75)}&\multicolumn{2}{c}{66.02($\uparrow$4.17)}&\multicolumn{2}{c}{66.15($\uparrow$3.48)}&&\multicolumn{2}{c}{61.06($\uparrow$2.37)}&\multicolumn{2}{c}{61.07($\uparrow$4.36)}&\multicolumn{2}{c}{60.98($\uparrow$4.67)}\\
				NST&\multicolumn{2}{c}{67.47($\uparrow$3.21)}&\multicolumn{2}{c}{66.56($\uparrow$4.71)}&\multicolumn{2}{c}{66.25($\uparrow$3.58)}&&\multicolumn{2}{c}{61.74($\uparrow$3.05)}&\multicolumn{2}{c}{61.25($\uparrow$4.54)}&\multicolumn{2}{c}{61.23($\uparrow$4.92)}\\
				VID&\multicolumn{2}{c}{67.59($\uparrow$3.33)}&\multicolumn{2}{c}{66.46($\uparrow$4.60)}&\multicolumn{2}{c}{66.51($\uparrow$3.84)}&&\multicolumn{2}{c}{61.93($\uparrow$3.24)}&\multicolumn{2}{c}{61.74($\uparrow$5.03)}&\multicolumn{2}{c}{61.93($\uparrow$5.62)}\\
				CRD&\multicolumn{2}{c}{67.88($\uparrow$3.62)}&\multicolumn{2}{c}{66.89($\uparrow$5.03)}&\multicolumn{2}{c}{66.63($\uparrow$3.96)}&&\multicolumn{2}{c}{61.86($\uparrow$3.17)}&\multicolumn{2}{c}{61.81($\uparrow$5.10)}&\multicolumn{2}{c}{61.82($\uparrow$5.51)}\\
				SemCKD&\multicolumn{2}{c}{67.91($\uparrow$3.65)}&\multicolumn{2}{c}{66.16($\uparrow$4.30)}&\multicolumn{2}{c}{66.78($\uparrow$4.11)}&&\multicolumn{2}{c}{62.22($\uparrow$3.53)}&\multicolumn{2}{c}{62.20($\uparrow$5.49)}&\multicolumn{2}{c}{62.05($\uparrow$5.74)}\\
				RKD&\multicolumn{2}{c}{66.98($\uparrow$2.72)}&\multicolumn{2}{c}{65.90($\uparrow$4.04)}&\multicolumn{2}{c}{66.09($\uparrow$3.42)}&&\multicolumn{2}{c}{61.78($\uparrow$3.09)}&\multicolumn{2}{c}{61.93($\uparrow$5.22)}&\multicolumn{2}{c}{61.81($\uparrow$5.50)}\\
				FSP&\multicolumn{2}{c}{67.43($\uparrow$3.17)}&\multicolumn{2}{c}{65.79($\uparrow$3.93)}&\multicolumn{2}{c}{66.41($\uparrow$3.74)}&&\multicolumn{2}{c}{61.63($\uparrow$2.94)}&\multicolumn{2}{c}{61.48($\uparrow$4.77)}&\multicolumn{2}{c}{61.42($\uparrow$5.11)}\\
				CC&\multicolumn{2}{c}{67.62($\uparrow$3.36)}&\multicolumn{2}{c}{66.54($\uparrow$4.68)}&\multicolumn{2}{c}{66.26($\uparrow$3.59)}&&\multicolumn{2}{c}{61.08($\uparrow$2.39)}&\multicolumn{2}{c}{61.79($\uparrow$5.08)}&\multicolumn{2}{c}{61.55($\uparrow$5.24)}\\
				\midrule
				Ours&\multicolumn{2}{c}{\textbf{68.25}($\uparrow$\textbf{3.99})}&\multicolumn{2}{c}{\textbf{67.03}($\uparrow$\textbf{5.17})}&\multicolumn{2}{c}{\textbf{67.09}($\uparrow$\textbf{4.42})}&&\multicolumn{2}{c}{\textbf{62.71}($\uparrow$\textbf{4.02})}&\multicolumn{2}{c}{\textbf{62.42}($\uparrow$\textbf{5.71})}&\multicolumn{2}{c}{\textbf{62.18}($\uparrow$\textbf{5.87})}\\
				\bottomrule
			\end{tabular}
		}
		\label{table:Imagenet}
	\end{table*}
	
	\textbf{Results on CIFAR-100 and CIFAR-10.} The results of the involved methods on the CIFAR-100 and CIFAR-10 testing sets are shown on the left and right of Table \ref{table:CIFAR}, respectively. The decoupling and analysis of the knowledge transfer process leads to certain performance improvements, as observed from the results. The proposed method achieves accuracy increments of 2.62\%, 2.35\%, and 4.20\% on the CIFAR-100 dataset, compared to the three baselines without CMKD, and 3.30\%, 5.43\%, and 4.24\% on the CIFAR-10 dataset. Moreover, the proposed method is the most effective in improving baseline accuracy, with a lead of 0.38\%, 0.69\%, and 0.69\% over sub-optimal methods with different backbones on the CIFAR-100 dataset. The corresponding metrics for the CIFAR-10 dataset are 0.50\%, 0.29\%, and 0.58\%. These results suggest that the DS-based structural knowledge defined by the proposed method exhibits superior performance compared to other approaches.
	
	\textbf{Results on STL-10 and Tiny-ImageNet.} The experimental results on the STL-10 and Tiny-ImageNet datasets are shown respectively on the left and right of Table \ref{table:StlImagenet}. Similar to previous experiments, the proposed method demonstrates the best knowledge transfer and performance improvement. Specifically, when ShuffleNets are utilized as the student backbone, the proposed method results in cross-modal accuracy improvements exceeding 10\% compared to the two versions of baselines. Across six sets of experimental results on the two benchmark datasets, the proposed method outperforms sub-optimal methods in terms of accuracy by 0.24\%, 0.36\%, and 0.46\% on STL-10, and by 0.48\%, 0.36\%, and 0.76\% on Tiny-ImageNet, respectively. These experimental results explicitly demonstrate the effectiveness of the proposed method over the transfer of informative dark knowledge.
	
	\textbf{Results on ImageNet-100 and ImageNet.} The left and right of Table \ref{table:Imagenet} show the results on ImageNet-100 and ImageNet datasets. From the experimental results, it can be seen that the proposed method achieves the best performance improvements on both datasets. On ImageNet-100 dataset, The proposed method improves the classification accuracy by 3.99\%, 5.17\%, and 4.42\% compared to the baseline, which are 0.34\%, 0.14\%, and 0.31\% improvements over the sub-optimal method for this dataset. Similar results can be seen in the experimental results on the ImageNet dataset shown on the right. The above experiments fully demonstrate the effectiveness of the proposed method and its robustness to the dataset.
	
	\subsection{Ablation Studies}
	The SRM, SEM and DCM are three core components of the proposed method, and the last two are also the keys to understanding and improving KD methods. In this part, ablation studies are performed to verify their respective effectiveness. Let the SRM denotes Eq. \eqref{srm2}, SEM-1 denotes the first term of Eq. \eqref{sem_revise} and SEM-2 for the second term of Eq. \eqref{sem_revise}, DCM-1 and DCM-2 for the first and second term of Eq. \eqref{dcm_revise}. Different models can be obtained for various combinations, and when the hyperparameters of these models are set to be the identical, a set of ablation studies can be obtained to verify the effectiveness of each module of the proposed method. Experimental results are shown in Table \ref{tab:ablation}.
	\begin{table}[!ht]
		\centering
		\caption{The ablation studies on the real-world CML dataset}
		\resizebox{0.98\columnwidth}{!}
		{
			\begin{tabular}{cccccc}
				\toprule
				SRM          & SEM-1   &SEM-2 &DCM-1 &DCM-2 &Acc(\%) \\
				\midrule
				\ding{52} &\ding{52} &\ding{52} &\ding{52} &\ding{52}  &\textbf{89.87}\\
				\midrule
				\ding{56} &\ding{52} &\ding{52} &\ding{52} &\ding{52}  &39.15\\
				\ding{52} &\ding{56} &\ding{52} &\ding{52} &\ding{52}  &86.75\\
				\ding{52} &\ding{52} &\ding{56} &\ding{52} &\ding{52}  &88.45\\
				\ding{52} &\ding{52} &\ding{52} &\ding{56} &\ding{52}  &88.95\\
				\ding{52} &\ding{52} &\ding{52} &\ding{52} &\ding{56}  &89.26\\
				\bottomrule
			\end{tabular}
		}
		\label{tab:ablation}
	\end{table}
	
	We choose the proposed real-world CML dataset to carry out the validation. It can be seen from the results that the highest classification accuracy of the CMKD is achieved when a total of five loss terms from three modules are used jointly. In addition, it can be seen that the SRM and SEM-1 have the largest and second largest effect on the classification performance.
	
	\subsection{Combining Different KD Objectives}
	To test the compatibility of the proposed method with other KD methods, we selected KD, Fitnets, AT, NST, CRD, RKD and test the performance of the proposed method in combination with them on the proposed real-world CML dataset to verify the plug-and-play capability of the proposed method. The backbone of teacher and student is set as ResNet32$\times$4 and ResNet8$\times$4, ShuffleNetV1. Experimental results are shown in Table \ref{table:KDcombine}. 
	\begin{table}[!ht]
		\centering
		\caption{The classification accuacy on the real-world CML dataset of combining different KD methods with the proposed one.}
		\resizebox{0.98\columnwidth}{!}
		{
			\begin{tabular}{lccccc}
				\toprule
				Teacher& \multicolumn{2}{c}{ResNet32$\times$4} &  \multicolumn{2}{c}{ResNet32$\times$4}\\
				Student& \multicolumn{2}{c}{ResNet8$\times$4} &  \multicolumn{2}{c}{ShuffleNetV1}\\
				\midrule
				Ours&89.87&-&88.02&-\\
				KD&88.29&-&86.45&-\\
				KD+Ours&90.32&($\uparrow$2.03)&88.84&($\uparrow$2.39)\\
				Fitnets&87.89&-&86.89&-\\
				Fitnets+KD&88.37&($\uparrow$0.48)&87.10&($\uparrow$0.21)\\
				Fitnets+Ours&89.93&($\uparrow$2.04)&88.31&($\uparrow$1.42)\\
				AT&87.53&-&87.16&-\\
				AT+KD&88.02&($\uparrow$0.49)&87.62&($\uparrow$0.46)\\
				AT+Ours&89.89&($\uparrow$2.36)&88.17&($\uparrow$1.01)\\
				NST&88.93&-&87.93&-\\
				NST+KD&89.15&($\uparrow$0.22)&88.14&($\uparrow$0.21)\\
				NST+Ours&90.12&($\uparrow$1.19)&88.55&($\uparrow$0.62)\\
				CRD&89.48&-&87.16&-\\
				CRD+KD&89.62&($\uparrow$0.14)&87.20&($\uparrow$0.04)\\
				CRD+Ours&90.24&($\uparrow$0.76)&88.19&($\uparrow$1.03)\\
				RKD&88.16&-&86.46&-\\
				RKD+KD&88.49&($\uparrow$0.33)&86.63&($\uparrow$0.17)\\
				RKD+Ours&90.17&($\uparrow$2.01)&88.34&($\uparrow$1.88)\\
				\bottomrule
			\end{tabular}
		}
		\label{table:KDcombine}
	\end{table}
	
	It can be seen from the results that both the introduction of KD and the proposed method can achieve performance improvements to the baseline method. However, comparing KD with the proposed method, it is obvious that the proposed method is more effective, which is reflected in the higher accuracy improvement caused by its integration. Therefore, the proposed method can be considered as a more efficient alternative to KD with wide adaptability and plug-and-play capability.
	
	\section{Conclusion}
	This paper finds a latent relationship between DS and feature discriminability. That is, the feature with higher discriminability should also has better structure at its dimensional aspect. Guided by this, we propose a novel CMKD method by functionally defining DS-based structured knowledge. Transferring the defined structured knowledge enforces the output and intermediate features to be channel-wise independent and uniformly distributed. This leverages the powerful model learned from the modality with high data quality to aid representation learning in the modality with low data quality, and effectively improves the performance of CML.
	
	Furthermore, we collect and open source a real-world CML dataset. It comprises over 10,000 paired optical and radar images from 8 categories, and can be used to demonstrate the effectiveness of CML and CMKD methods. The experimental results verify that our proposed method achieves superior CML performance in different datasets and settings.
	
	While some progress has been achieved, shortcomings still exist, which is mainly reflected in the fact that the difficulties in the construction of paired CML data have not been given sufficient attention at the algorithm design aspect. In future work, we aim to address this issue through self-supervised KD. Moreover, several unresolved issues in the CML community, including the learning with point cloud, infrared, audio and video data are all issues we are considering.
	
	\bibliographystyle{spbasic} 
	\bibliography{myreffull}
\end{document}